\documentclass[10pt,twocolumn,letterpaper]{article}

\PassOptionsToPackage{table}{xcolor}
\usepackage[pagenumbers]{wacv}    %
\usepackage[T1]{fontenc}

\usepackage{multirow}
\usepackage{makecell}
\usepackage{mathtools}
\usepackage{pifont}
\usepackage{amsthm}
\usepackage{thmtools}
\usepackage{thm-restate}
\usepackage{tabularx}
\usepackage[normalem]{ulem}
\usepackage{listings}
\usepackage{algorithm}
\usepackage{xcolor}

\usepackage{amsmath,amsfonts,bm}

\def\eqref#1{equation~\ref{#1}}

\def\1{\bm{1}}

\def\ve{{\bm{e}}}

\def\vm{{\bm{m}}}

\def\vq{{\bm{q}}}

\def\mQ{{\bm{Q}}}

\def\mW{{\bm{W}}}

\DeclareMathAlphabet{\mathsfit}{\encodingdefault}{\sfdefault}{m}{sl}
\SetMathAlphabet{\mathsfit}{bold}{\encodingdefault}{\sfdefault}{bx}{n}

\def\gB{{\mathcal{B}}}
\def\gC{{\mathcal{C}}}
\def\gD{{\mathcal{D}}}
\def\gE{{\mathcal{E}}}

\def\gL{{\mathcal{L}}}
\def\gM{{\mathcal{M}}}

\def\gS{{\mathcal{S}}}
\def\gT{{\mathcal{T}}}

\def\sR{{\mathbb{R}}}

\definecolor{wacvblue}{rgb}{0.21,0.49,0.74}
\usepackage[pagebackref,breaklinks,colorlinks,allcolors=wacvblue]{hyperref}
\usepackage[accsupp]{axessibility}  %

\def\wacvPaperID{1049} %
\def\confName{WACV}
\def\confYear{2027}

\definecolor{LightBlue}{rgb}{0.85, 0.90, 0.94}
\definecolor{codeblue}{rgb}{0.25,0.5,0.5}
\definecolor{codegreen}{rgb}{0,0.6,0}
\definecolor{codekw}{rgb}{0.85, 0.18, 0.50}

\def\eqref#1{Eq.~(\ref{#1})}

\newcommand{\best}[1]{\textbf{#1}}
\newcommand{\second}[1]{\underline{#1}}
\newcommand{\tick}{\ding{51}}

\newcommand{\xmark}{\textcolor{gray}{\ding{55}}}
\newcommand{\cg}{\cellcolor{LightBlue}}

\newcommand\Mark[1]{\textsuperscript#1}

\title{Align Your Query: \\Representation Alignment for Multimodality Medical Object Detection}

\author{
Ara Seo\Mark{{1,2}}\thanks{Equal contribution.} \qquad Bryan Sangwoo Kim\Mark{{1,3}}\footnotemark[1] \qquad Hyungjin Chung\Mark{{4}} \qquad Jong Chul Ye\Mark{{1}} \\
\Mark{1}KAIST \quad \Mark{2}Hanwha Systems \quad \Mark{3}Stanford University \quad \Mark{4}Korea University }

\begin{document}
\maketitle

\begin{abstract}

Medical object detection suffers when a single detector is trained on mixed medical modalities (\textit{e.g.}, CXR, CT, MRI) due to heterogeneous statistics and disjoint representation spaces. To address this challenge, we turn to representation alignment, an approach that has proven effective for bringing features from different sources into a shared space. Specifically, we target the representations of DETR-style object queries and propose a simple, detector-agnostic framework to align them with modality context.
First, we define \textit{modality tokens}: compact, text-derived embeddings encoding imaging modality that are lightweight and require no extra annotations. We integrate the modality tokens into the detection process via \textit{Multimodality Context Attention} (MoCA), mixing object-query representations via self-attention to propagate modality context within the query set. This preserves DETR-style architectures and adds negligible latency while injecting modality cues into object queries.
We further introduce \textit{QueryREPA}, a short pretraining stage that aligns query representations to their modality tokens using a task-specific contrastive objective with modality-balanced batches.
Together, MoCA and QueryREPA produce modality-aware, class-faithful queries that transfer effectively to downstream training. Across diverse modalities trained altogether, the proposed approach consistently improves AP with minimal overhead and no architectural modifications, offering a practical path toward robust multimodality medical object detection.
\end{abstract}

\section{Introduction}
\label{intro}

Medical object detection, the task of identifying and localizing specific anatomical structures or abnormalities within medical images, is a cornerstone of modern computer-aided diagnosis and clinical decision support systems. Conventional object detection methods originally developed for natural images~\cite{sun2021sparse, zhu2020deformable, zhang2022dino, li2022grounded, chen2023diffusiondet, liu2024grounding} achieve strong results in the medical domain when applied within a single imaging modality. However, the prevailing trend in foundation models is towards the integration of diverse data types, and the biomedical domain is no exception~\cite{moor2023foundation, tu2024towards}. This paradigm introduces significant complexity in the biomedical field, which is characterized by a wide array of imaging modalities such as X-ray, CT, MRI, colonoscopy, etc.

This inherent heterogeneity in medical data of different imaging modalities often leads to degraded performance when a single model is trained on a mixed multimodality\footnote{\textit{Multimodality} is defined to be the state of containing multiple imaging modalities (\textit{e.g.}, CXR, CT, MRI); this is not to be confused with \textit{multimodal}, the state of containing multiple data types (\textit{e.g.}, text, image).} dataset~\cite{guan2021domain, kang2023structure, yang2024cross}. The distinct statistical properties and visual characteristics of each imaging type create a complex, disjoint representation space~\cite{cho2025medisyngeneralisttextguidedlatent}.
One response to this heterogeneity is to avoid it altogether: because the imaging modality is normally recorded in acquisition metadata, each image could simply be routed to a modality-specific expert detector. However, such specialization complicates deployment in proportion to the number of modalities and forfeits any benefit one modality could lend another. Thus, in this work we aim to retain the deployment simplicity and cross-modality transfer of a unified detector while removing the degradation that motivates specialization (Sec.~\ref{sec:comparison}).
The central problem, therefore, is the generation of efficient and effective representations that can generalize across a mixed corpus of highly diverse medical imaging data.

To address this, we turn to \textit{representation alignment}, a technique that has recently gained significant traction for its ability to harmonize representations from disparate sources~\cite{wang2020understanding}. Recently popularized in generative models through methods like REPA~\cite{yu2024representation}, representation alignment seeks to bridge the semantic gap between different modalities, thereby creating a more cohesive and informative shared representation space~\cite{lee2025aligning, yoon2025visual}. We posit that this concept can be repurposed for the task of medical object detection in multimodality datasets.
Specifically, we introduce a novel approach that redefines the role of representation alignment by focusing on \emph{object queries}, the learnable embeddings that directly inform class prediction and bounding box regression in modern detection architectures~\cite{carion2020end}.
We align the object queries with \textit{modality tokens}, which serve as text-derived representations that encode each imaging modality (\textit{e.g.}, CXR, CT, MRI, endoscopy) and target class.
The modality tokens act as stable anchors shared across the mixed multimodality dataset with favorable properties: they are lightweight to generate, require no extra annotations, and make modality context explicit without altering backbone or head designs.

For integration of modality tokens, we propose \textbf{M}ultim\textbf{o}dality \textbf{C}ontext \allowbreak\textbf{A}ttention (\textbf{MoCA}), a novel self-attention mechanism designed to mix the representations of object queries.
Rather than adding a separate language stream, MoCA simply appends the relevant modality token to the detector’s query set and lets the decoder’s self-attention mix the token with object queries. This keeps the architecture faithful to DETR-style designs, adds negligible latency, and injects modality cues within the evolving query representations precisely where decisions are formed.
Inspired by recent methods for multimodal integration in generative models~\cite{esser2024scaling}, MoCA is, to the best of our knowledge, the first method to explore \textit{intra-query mixing} of representations for object detection.
This represents an important departure from conventional multimodal integration techniques that pre-train the image encoder or utilize cross-attention~\cite{li2022grounded, liu2024grounding}, widening our interest to multimodal integration of object queries.

Furthermore, to strengthen the alignment, we introduce a pre-training stage where object query representations are aligned with modality tokens guided by a contrastive \textbf{Query Representation Alignment} (\textbf{QueryREPA}) loss.
This alignment occurs before any detection head is trained, shaping a query manifold that is both modality-aware and class-faithful. To facilitate this process and prevent any single modality from dominating the training, we perform \textit{modality batch sampling} to ensure each batch contains a balanced mix of images from different modalities.

\begin{figure*}[t]
    \centering
    \includegraphics[width=.9\linewidth]{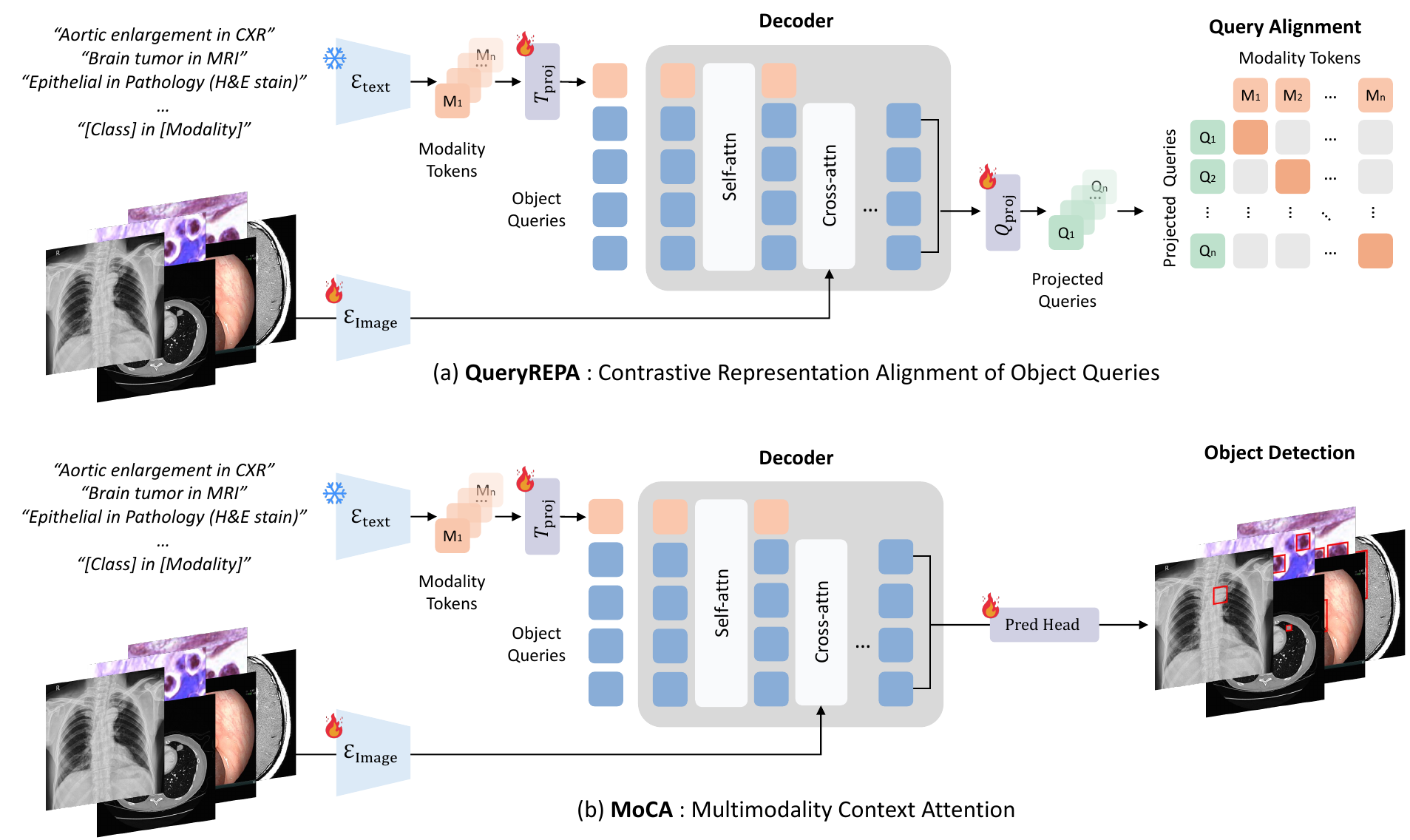}
    \caption{\textbf{(a) Contrastive Representation Alignment of Object Queries (QueryREPA):} During a pretraining stage, each image in a modality-balanced batch is associated with a text-derived \textit{modality token}. For each image, we mean-pool the decoder’s object-query embeddings into a single query representation, project it into the modality-token space, and optimize a contrastive alignment loss between the projected queries and the modality tokens, yielding a modality-aware query manifold.
    \textbf{(b) Multimodality Context Attention (MoCA):}
    For a given image, a modality token is selected and concatenated with the set of object queries to form an augmented query set. Information fusion occurs in the self-attention layer of the decoder, where each query attends to the modality token to attain modality-specific context.}
    \label{fig:main}
\end{figure*}

In summary, our main contributions are:
\begin{enumerate}
    \item A simple, detector-agnostic formulation of modality tokens that makes modality and class context explicit with minimal overhead. These modality tokens are lightweight to generate and require no extra annotations.
    \item \textit{Multimodality Context Attention (MoCA)}, a drop-in self-attention mechanism that appends modality tokens to queries, strengthening multimodality detection without architectural modifications.
    \item \textit{QueryREPA}, a pretraining stage that aligns query representations to modality tokens using a task-specific contrastive loss. Combined with MoCA, QueryREPA enables usage of large scale unlabeled datasets and effectively boosts downstream AP across diverse medical imaging modalities.
\end{enumerate}

\section{Related Work}
\label{related-work}

\subsection{Multimodality in the Medical Domain}

The rapid push toward generalist biomedical AI has intensified interest in training models across heterogeneous imaging modalities (e.g., CXR, CT, MRI, pathology), but domain shifts and modality-specific statistics often degrade performance in unified settings \cite{moor2023foundation,tu2024towards,guan2021domain,kang2023structure,yang2024cross}.
In parallel, contrastive pretraining with paired medical image-text data (e.g., ConVIRT~\cite{zhang2022contrastive}) demonstrates that textual supervision can structure representation spaces with clinically meaningful semantics.
More recently, M$^3$Bind~\cite{liu2025multimodal} proposes a pretraining framework that uses a shared text embedding space to bind diverse medical imaging modalities (\textit{e.g.}, X-ray, CT, retina) without requiring explicit cross-modality image pairs.
Together these motivate compact, text-derived signals as efficient carriers of modality context, which we introduce at the level of object queries.

\subsection{Object Detection}

Transformer-based detectors first introduced in DETR~\cite{carion2020end} have reshaped end-to-end detection through query-based decoding and set prediction, with subsequent advances improving optimization (\textit{e.g.}, Deformable DETR~\cite{zhu2020deformable}) and feature sampling (\textit{e.g.}, DINO~\cite{zhang2022dino}), and proposal design (\textit{e.g.}, Sparse R-CNN~\cite{sun2021sparse}, DiffusionDet~\cite{chen2023diffusiondet}).
For language-guided open-set detection, GLIP~\cite{li2022grounded} and Grounding DINO~\cite{liu2024grounding} blend text with vision, typically by threading text tokens via cross-attention into the decoder.

In the medical setting, recent efforts have explored DETR-style pipelines for medical object detection~\cite{ickler2023taming}. However, these models are not self-configuring and therefore require dataset-specific tuning, limiting their applicability across heterogeneous medical datasets.

\subsection{Representation Alignment for Object Detection}

Representation alignment reconciles heterogeneous embeddings, improving transfer and robustness~\cite{wang2020understanding,duan2022multi,xiao2024targeted}.
For \textit{object detection}, such methods can be grouped into three paradigms:
\textbf{(i) backbone-level alignment} of global image features, where contrastive or CLIP-style objectives act only on outputs of the \textit{image encoder} and do not modify the logic of the detection model at all (\textit{e.g.}, MoCo~\cite{he2020momentum}, SimCLR~\cite{chen2020simple}, SwAV~\cite{caron2020unsupervised}, ConVIRT~\cite{zhang2022contrastive}); \textbf{(ii) region-centric alignment}, which aligns \textit{local} pre-defined region or proposal features to text descriptions (\textit{e.g.}, RegionCLIP~\cite{zhong2022regionclip}, DITO~\cite{kim2024region}); and \textbf{(iii) query-level alignment (ours)}, which directly aligns object-level embeddings or ``queries'' to compact modality tokens inside the decoder. Therefore, our framework differs from existing works in both what is being aligned \textit{and} the target of the alignment.
The first two paradigms are complementary to ours: backbone-level pretraining could precede our query-level alignment.

\section{Representation Alignment for Medical Object Detection}
\label{method}

We address multimodality heterogeneity by making modality context explicit inside the query set of detectors. First, we construct compact, text-derived \textbf{modality tokens} for each imaging modality, optionally conditioned on a target class. Next, \textbf{Multimodality Context Attention (MoCA)} appends the relevant token to the object-query set and mixes it with queries via decoder self-attention, adding negligible latency (Fig.~\ref{fig:main}b). 
Finally, \textbf{Query Representation Alignment (QueryREPA)} pretrains the query space by aligning query statistics to their modality tokens using a contrastive objective with modality-balanced batches (Fig.~\ref{fig:main}a).

\subsection{Modality Tokens}

\paragraph{Construction of Modality Tokens.}
Each modality token contains semantic information regarding its class and imaging modality, encouraging disentanglement among imaging modalities in multimodality datasets. Specifically, let $\gT_{s} = \{\texttt{[CLS]}, w_1, w_2, \cdots, w_L, \texttt{[SEP]}\}$ be a sequence of words comprising the textual input
describing a context $s\in\gS$, where $\gS=\gD$ in the \textit{modality-only} configuration (indexing imaging modality $d\in\gD$; \textit{e.g.}, CXR, CT, MRI, pathology, colonoscopy) and $\gS=\gD\times\gC$ in the \textit{class-conditioned} configuration (additionally indexing a target class $c\in\gC$; \textit{e.g.}, aortic enlargement).
A pretrained text encoder $\gE$ transforms $\gT_{s}$ into a set of embeddings:
\begin{equation}
    \gE(\gT_{s}) = [\ve_{\texttt{[CLS]}}, \ve_1, \ve_2, \cdots, \ve_L, \ve_{\texttt{[SEP]}}] \in \sR^{d_\text{text} \times (L+2)}
    \label{eq:1}
\end{equation}
where $d_\text{text}$ is the dimension of the text-encoder output space.
Rather than carrying all $L+2$ embeddings, we use only the \texttt{[CLS]} embedding $\ve_{\texttt{[CLS]}} \in \sR^{d_\text{text} \times 1}$, which summarizes the sequence in a single vector.
The \texttt{[CLS]} token is mapped by the linear projection $\mW \in \sR^{d_\text{model} \times d_\text{text}}$ to define the modality token $\vm_{s}$ generated from $\gT_{s}$:
\begin{equation}
    \vm_{s} := \mW\ve_{\texttt{[CLS]}} \in \sR^{d_\text{model}}
\end{equation}
We create a modality token $\vm_{s}$ for each context $s\in\gS$ to create a fixed set of modality tokens $\gM_\gE$:
\begin{equation}
    \gM_\gE := \{ \vm_{s} : {s \in \gS} \} .
    \label{eq:modality-tokens}
\end{equation}
In the modality-only configuration, $s$ is read from the DICOM Modality tag or fixed by the acquisition protocol, requiring no annotation. The class-conditioned configuration is designed for cases where a specific finding is sought in advance, mirroring the indication recorded on an imaging requisition and the interaction premise of text-promptable medical models~\cite{killeen2025fluorosam, liu2025medsam3}; in our experiments the class component is taken from dataset annotations, and we discuss the implications in Sec.~\ref{sec:comparison}.
Furthermore, different encoders $\gE$ yield distinct token sets $\gM_\gE$, and our framework improves performance regardless of the choice (see Appendix~\ref{sec:appendix-encoders}).

\subsection{MoCA : Multimodality Context Attention}
 
\begin{figure*}[t]
    \centering
    \includegraphics[width=\linewidth]{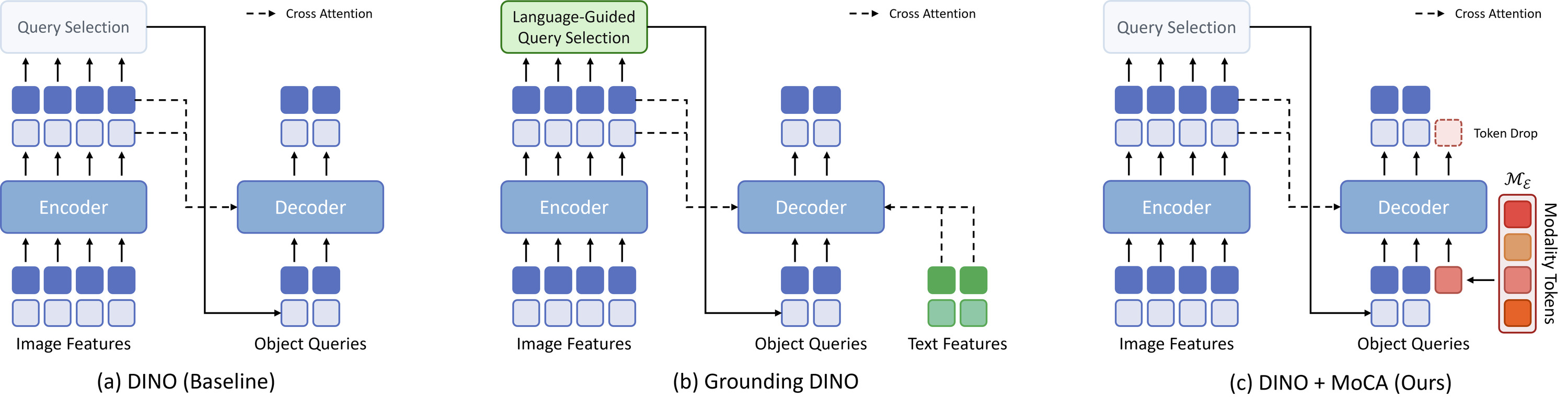}
    \caption{
    Comparison of text/modality integration in DETR-style detectors. \textbf{(a) DINO (baseline):} image-only encoder–decoder with query selection. \textbf{(b) Grounding DINO:} language-guided query selection and cross-attention in the decoder (object queries attend to a sequence of text tokens). \textbf{(c) Ours (DINO+MoCA):} append compact modality tokens to the object queries and fuse by self-attention in the decoder; the tokens act as semantic anchors that refine queries in a modality-aware way with minimal overhead.
    }
    \label{fig:comparison}
\end{figure*}

To integrate the knowledge of modality tokens $\gM_\gE$, we propose MoCA, a novel method to refine object queries during the detection process via self-attention in the detection decoder.
Contrary to prior methods that do not exhibit multimodal integration (Fig.~\ref{fig:comparison}a) or integrate via cross attention (Fig.~\ref{fig:comparison}b), we create a fused representation space between modality tokens and object queries (Fig.~\ref{fig:comparison}c).

Formally, given an input image with context $s\in\gS$,
we leverage the modality token $\vm_{s} \in \gM_\gE$ as the target for fusion.
The matrix of $N$ object queries $\mQ=[\vq_1, \cdots, \vq_N] \in \mathbb{R}^{N \times d_{model}}$ is concatenated with the modality token $\vm_{s}$ through a learnable linear projection $f_\theta$ to form an augmented query set $\tilde{\mQ}$ as follows:
\begin{equation}
    \tilde{\mQ}=[\vq_1, \cdots, \vq_N, f_\theta{(\vm_{s})}] \in \mathbb{R}^{(N+1)\times d_{model}}
\end{equation}

This augmented set $\tilde{\mQ}=[\mQ,f_\theta{(\vm_{s})}]$ is then fed into a Multi-head Self-Attention (MSA) layer of the detection decoder.
Let $\mQ^{(l)}$ denote the set of queries $\mQ$ at layer $l$ of the detection decoder.
The overall attention operation is,
\begin{equation}
    \text{MSA}(\tilde{\mQ}) = \text{softmax} \left( \frac{(\tilde{\mQ}\mW_Q)(\tilde{\mQ}\mW_K)^T}{\sqrt{d_k}} \right)(\tilde{\mQ}\mW_V)
\end{equation}
\begin{equation}
    \mQ^{(l+1)} = \text{MSA}\left(\tilde{\mQ}^{(l)}\right)_{1:N} + \mQ^{(l)}
\end{equation}
where $\mW_Q, \mW_K, \mW_V$ are learnable projection matrices,
and the modality-token row is discarded before the residual connection.
Because $\mQ^{(l+1)}$ is an explicitly learned function of $(\mQ^{(l)},\vm_{s})$, training with the final detection loss encourages the model to \emph{use} the informative pathway from $\vm_{s}$. This increases statistical dependence between $\mQ^{(l+1)}$ and $\vm_{s}$, raising the mutual information $I(\mQ^{(l+1)};\vm_{s})$ relative to an identical decoder lacking $\vm_{s}$.

The critical consequence of this formulation is that every query in the augmented set can attend to every other query, which enables the $N$ object queries to \textit{refer to} information from the modality token.
This process is repeated across all decoder layers, allowing for iterative refinement of object queries based on the infusion of modality-specific context.

\paragraph{Modality tokens at inference.}
Since $\gM_\gE$ is a fixed set of tokens, all tokens are precomputed once with the frozen encoder $\gE$ and cached, so $\gE$ is never instantiated at inference and MoCA adds only a single token to the query set (Tab.~\ref{tab:decoder_latency}).

\subsection{QueryREPA: Contrastive Representation Alignment of Object Queries}

Modern query-based detectors use an encoder–decoder transformer that maintains, at each decoder layer $l$, a set of object-query embeddings $\{\vq_i^{(l)}\}_{i=1}^N \in \sR^{d_\text{model}}$. In standard training, these queries are initialized identically across all imaging modalities and are shaped only indirectly through the detection loss.
As a result, they do not explicitly encode modality-specific semantics and often form an entangled representation space across modalities (Fig.~\ref{fig:umap-queryrepa}, left).
QueryREPA takes the decoder outputs $\{\vq_i^{(l)}\}$ at a chosen layer $l$ and aligns them with their corresponding modality tokens. We apply QueryREPA only to layers after cross-attention has injected image features, ensuring that the queries we align are informative about the input image.

\paragraph{Query Representation Alignment.}

For the set of query embeddings {$\{\vq_i^{(l)}\}_{i=1}^N$}, we form a mean query representation $\overline{\vq}^{(l)}=\frac{1}{N}\sum_{i=1}^N{\vq_i^{(l)}}$,
which we align to the modality token $\vm_{s} \in \gM_\gE$.
A learnable projection $g_\phi$ maps $\overline{\vq}^{(l)}$ into the latent space of $\gM_\gE$.
The Query Representation Alignment (\textbf{QueryREPA}) loss is defined as:
\begin{equation}
    \mathcal{L}_\text{QRA}(\overline{\vq}^{(l)}, \vm_{s}) = -\log\frac{\exp(\text{sim}(g_\phi(\overline{\vq}^{(l)}), \vm_{s})/\tau)}{\sum_j{\exp(\text{sim}(g_\phi(\overline{\vq}^{(l)}), \vm_{s}^{(j)})/\tau)}}
    \label{eq:queryrepa}
\end{equation}
where $\text{sim}(\cdot,\cdot)$ denotes cosine similarity and $\tau>0$ denotes the temperature. This loss is a direct adaptation of the InfoNCE loss~\cite{oord2018representation} well established in the context of representation learning.
We run QueryREPA as a short pretraining stage before end-to-end detection training: the backbone and decoder are updated using only $\mathcal{L}_\text{QRA}$, with the detection heads unused. The full detector is then trained with standard detection losses, initialized from the pretrained backbone and decoder, with the heads randomly initialized.
Fig.~\ref{fig:umap-queryrepa} shows the effect: the mean query representations $\overline{\vq}^{(l)}$ overlap substantially across modalities before pretraining and separate into disentangled clusters afterwards, with the silhouette score rising from 0.489 to 0.747.

\paragraph{Modality Batch Sampling.}
\label{modality_batch_sampling}

Let $\gD$ denote the set of imaging modalities (e.g., CXR, CT, MRI, endoscopy, pathology), $|\gD|=M$.
Let $\gB=\left\{(x_b,s_b)\right\}_{b=1}^{B}$, and let $d(s)$ denote the modality component of context $s$. We construct mini-batches such that all samples come from distinct modalities:
\begin{equation}
    d(s_b) \neq d(s_{b'}) \quad \text{for all} \ b \neq b', \quad B \leq M.
    \label{eq:8}
\end{equation}

In practice, we implement a round-robin, per-modality queue sampler that (i) cycles uniformly over $\gD$ to mitigate modality imbalance, (ii) draws one example per selected modality, and (iii) optionally shuffles the class component of $s_b$ within each modality-specific queue. This guarantees that, for any anchor image $x$ from modality $d$, the in-batch \emph{negatives} used by the QueryREPA loss of \eqref{eq:queryrepa} are tokens from \emph{other} modalities $d'\neq d$.
Such batch construction focuses the contrast explicitly on \emph{cross-modality} separability, yielding stronger, well-conditioned gradients for modality-aware alignment. When $B<M$, we randomly sample the subset of covered modalities with uniform probability across iterations so that, amortized over training, each modality pair is contrasted with high probability.

\paragraph{QueryREPA as MI maximization.}

The proposed QueryREPA induces mutual information maximization of object queries.
Define $U^{(l)} := g_\phi\big(\overline{\vq}^{(l)}\big)$ the projected query statistic at decoder layer $l>1$ and $V:= \vm_{s}$ the modality token.
Using the InfoNCE objective on positive pairs $(U^{(l)}, V)$ and $K$ in-batch negatives $\{ V_k \}$, $\mathcal{L}_\text{QRA}^{(l)}$ can be represented as follows:
\begin{equation}
    \resizebox{\columnwidth}{!}{$\displaystyle
    \gL_\text{QRA}^{(l)} = -\mathbb{E}\left[ \log{\frac{\exp(\text{sim}(U^{(l)}, V)/\tau)}{\exp(\text{sim}(U^{(l)}, V)/\tau) + \sum_{k=1}^K{\exp(\text{sim}(U^{(l)}, V_k)/\tau)}}} \right]$}.
    \label{eq:infonce}
\end{equation}

\begin{restatable}{remarkstyle}{lowerbound}
    \label{remark:lowerbound}
    Given the InfoNCE objective of \eqref{eq:infonce} on positive pairs $(U^{(l)}, V)$ and $K$ in-batch negatives $\{ V_k \}$, the standard lower bound is expressed as
    \begin{equation}
        I(U^{(l)};V) \geq \log(1+K) - \gL_\text{QRA}^{(l)}.
        \label{eq:lowerbound}
    \end{equation}
\end{restatable}

Thus, minimizing $\gL_\text{QRA}$ can be interpreted as maximizing a variational lower bound on $I\big(g_\phi(\overline{\vq}^{(l)});\vm_{s}\big)$.
Because our modality batch sampling ensures $V_k$ are tokens from different modalities, the same objective decreases the association of the queries with mismatched modalities (i.e., pushes $I\big(U^{(l)};V_k\big)$ down via the softmax competition). The bound tightens as $K$ increases, indicating that a larger $B$ (up to $M$) would result in higher performance.

\section{Experiments}

\subsection{Experimental Settings}

\paragraph{Datasets.} 

We create a \textit{single mixed multimodality dataset} via aggregation of multiple sub-datasets. The sub-datasets we use vary in terms of image resolution, number of annotations, and class distributions, to ensure a robust evaluation across different imaging modalities.
Specifically, we use VinBigData Chest X-ray~\cite{vinbig} for CXR, COVID-19 CT~\cite{covid19ct} and LIDC-IDRI~\cite{lidc} for CT, NeoPolyp~\cite{neopolyp} for colonoscopy, BR35H brain tumor~\cite{br35h} and ACDC cardiac MRI~\cite{acdc} for MRI, MoNuSeg~\cite{monuseg} for pathology.
Detailed description regarding the datasets used is provided in Appendix~\ref{sec:appendix-dataset}.

\paragraph{Training Details.}
We train all models on a single RTX~3090/4090 GPU and report results from the checkpoint achieving the highest validation AP.
Models are optimized using AdamW; DINO uses a base learning rate of $2{\times}10^{-4}$ for 36 epochs with MultiStep decay at epoch 30, and Deformable DETR $1{\times}10^{-4}$ for 50 epochs with decay at epoch 40.
We adopt standard multi-scale data augmentation. Detailed configurations, augmentation schedules, and hyperparameters are provided in Appendix~\ref{sec:training-details}.

We assess the impact of using various encoders (\textit{i.e.}, CLIP~\cite{radford2021learning}, BiomedCLIP~\cite{zhang2023biomedclip}, and PubMedCLIP~\cite{eslami2023pubmedclip}) for the construction of modality tokens. BiomedCLIP and PubMedCLIP are variants of CLIP, tuned specifically for the medical domain. For the image encoder, all experiments are conducted using the ResNet-50 image encoder across all baseline detectors for fair comparison.

\paragraph{Evaluation metrics.}

We evaluate using the standard Average Precision (AP) metric (averaged over multiple IoU thresholds) and AP$_{50}$ (at an IoU threshold of 0.5).
For comparison with state-of-the-art detection methods, we additionally provide metrics AP$_{75}$, AP$_{s}$, AP$_{m}$, AP$_{l}$.

\begin{table}[t]
\caption{Quantitative comparison with state-of-the-art methods in medical object detection. Applying \colorbox{LightBlue}{our method} (MoCA + QueryREPA) on the DINO~\cite{zhang2022dino} detector with $\gE$ as CLIP, in both the modality-only ($\gS=\gD$) and class-conditioned ($\gS=\gD\times\gC$) configurations, outperforms all other detection methods. \textbf{Bold} = best, \underline{underline} = second best.}
\label{tab:2}
\begin{center}
\resizebox{\linewidth}{!}{%
\begin{tabular}{l|c|c|c|c|c|c}
\toprule
Method & AP & AP$_{50}$ & AP$_{75}$ & AP$_s$ & AP$_m$ & AP$_l$ \\
\midrule
Deformable DETR~\cite{zhu2020deformable} \quad &
35.0 & 54.4 & 37.8 & 17.4 & 27.6 & 38.2 \\
Sparse R-CNN~\cite{sun2021sparse} &
35.9 & 54.4 & 38.8 & 18.4 & 29.0 & 38.0 \\
DiffusionDet~\cite{chen2023diffusiondet} &
38.2 & 60.3 & 40.1 & 22.4 & 30.6 & 41.6 \\
GLIP~\cite{li2022grounded} &
37.4 & 57.5 & 39.9 & 20.8 & 30.5 & 42.2 \\
GLIP-A~\cite{li2022grounded} &
37.7 & 58.3 & 40.1 & 20.9 & 31.0 & 41.4 \\
GLIP-B~\cite{li2022grounded} &
37.1 & 57.4 & 39.4 & 22.4 & 29.7 & 40.8 \\
GLIP-C~\cite{li2022grounded} &
37.4 & 57.5 & 40.3 & 21.3 & 29.9 & 42.0 \\
Grounding DINO~\cite{liu2024grounding} &
32.7 & 50.9 & 35.1 & 17.9 & 26.2 & 33.9 \\
RTMDet~\cite{lyu2022rtmdet} &
{38.5} & {57.4} & {41.9} & {22.4} & {32.0} & {42.5} \\
{Co-DETR~\cite{zong2023detrs}} &
{40.2} & {59.1} & {\textbf{43.6}} & {\second{25.9}} & {32.7} & {43.3} \\
DINO~\cite{zhang2022dino} &
37.7 & 58.6 & 37.8 & 19.9 & 29.8 & 41.7 \\

\rowcolor{LightBlue}
+ Ours ($\gS=\gD$) &
\second{40.7} & \second{65.1} & 42.7 & \best{26.1} & \second{33.0} & \second{44.2} \\
\rowcolor{LightBlue}
+ Ours ($\gS=\gD\times\gC$) &
\best{41.1} & \best{65.5} & \second{43.4} & 25.4 & \best{34.6} & \best{44.9} \\

\bottomrule
\end{tabular}
}%
\end{center}
\end{table}

\begin{table}[t]
\centering
\caption{Results for alternative token designs. \textbf{Bold} = best, \underline{underline} = second best.}
\label{tab:modality-tokens}
\setlength{\tabcolsep}{3pt}
\resizebox{\linewidth}{!}{
\begin{tabular}{llccccccc}
\toprule
Token & Partition & AP & AP$_{50}$ & AP$_{75}$ & AP$_s$ & AP$_m$ & AP$_l$ \\
\midrule
Baseline (no token) & --- & 37.7 & 58.6 & 37.8 & 19.9 & 29.8 & 41.7 \\
\midrule
\multicolumn{8}{l}{\textit{Capacity controls}} \\
Learnable & global & 37.6 & 58.1 & 40.6 & 21.4 & 29.6 & 41.7 \\
Random    & modality & 37.9 & 58.1 & 40.8 & 19.9 & 30.3 & 41.8 \\
\midrule
\multicolumn{8}{l}{\textit{Correctness control}} \\
Shuffled  & modality & 38.3 & 59.0 & 40.9 & 20.7 & 30.3 & 42.3 \\
\midrule
\multicolumn{8}{l}{\textit{Dataset-source control (text-derived)}} \\
Per-dataset & dataset & 38.0 & 58.4 & 40.8 & 21.0 & 29.4 & 41.4 \\
\midrule
\multicolumn{8}{l}{\textit{Modality and class semantics (text-derived)}} \\
\rowcolor{LightBlue}
\texttt{\{MODALITY\}} & modality & 40.7 & 65.1 & 42.7 & \underline{26.1} & 33.0 & 44.2 \\
\rowcolor{LightBlue}
\texttt{\{CLASS\}}    & class   & \underline{41.0} & \underline{65.4} & \underline{43.2} & \textbf{27.6} & \underline{33.6} & \underline{44.5} \\
\rowcolor{LightBlue}
\texttt{\{CLASS in MODALITY\}} & class & \textbf{41.1} & \textbf{65.5} & \textbf{43.4} & 25.4 & \textbf{34.6} & \textbf{44.9} \\
\bottomrule
\end{tabular}
}
\end{table}

\subsection{Comparison Results}
\label{sec:comparison}

\paragraph{Quantitative Results.}
Tab.~\ref{tab:2} reports results on the mixed multimodality medical dataset. Built on DINO, our method raises AP from 37.7 to 41.1 and AP$_{50}$ from 58.6 to 65.5, outperforming prior detectors in overall AP while remaining strong across object scales. The gains over language-guided detectors such as GLIP and Grounding DINO suggest that appending compact modality tokens to the query set is more effective here than threading text through cross-attention. Notably, the annotation-free modality-only configuration ($\gS=\gD$) already reaches 40.7 AP, exceeding competing methods without any class information; the class-conditioned configuration adds a further 0.4 AP. Appendix~\ref{sec:appendix-encoders} shows the pattern holds across CLIP, BiomedCLIP, and PubMedCLIP, and also for Deformable DETR, confirming the robustness of our framework.

\begin{figure}[t]
    \centering
    \includegraphics[width=\linewidth]{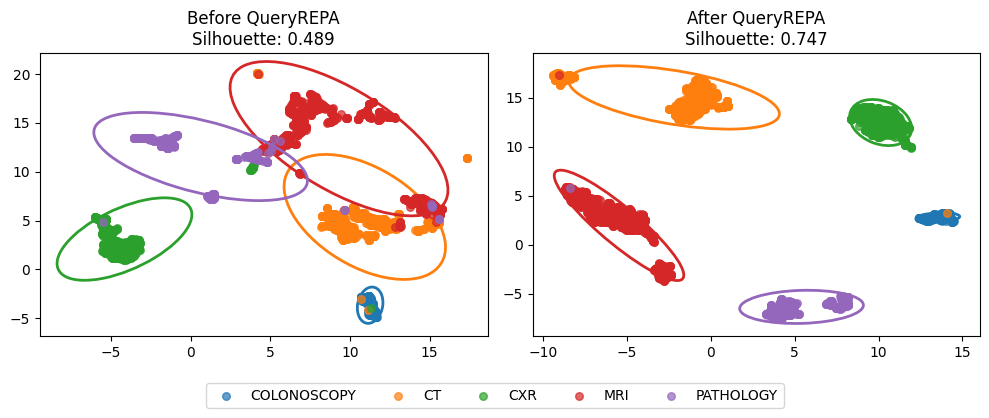}
    \caption{
        \textbf{QueryREPA incorporates modality-aware semantics into the query space.}
        UMAP embeddings of the mean query representation $\bar{q}^{(l)}$, computed per image and colored by imaging modality. Before QueryREPA (left) the representations are highly entangled across modalities, whereas after QueryREPA (right) they separate into disentangled clusters.
        }
    \label{fig:umap-queryrepa}
\end{figure}

\begin{figure*}[t]
    \centering
    \includegraphics[width=.95\linewidth]{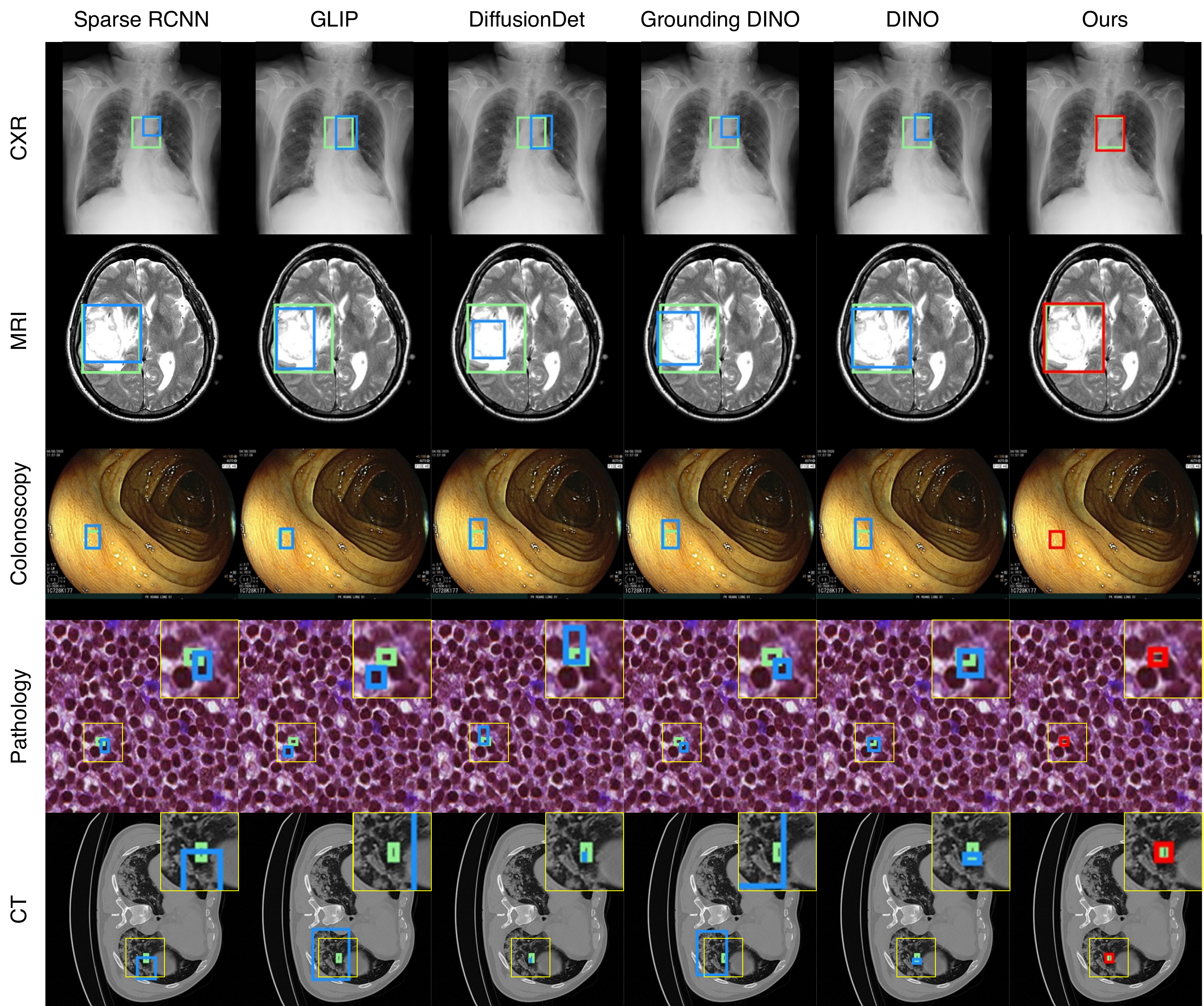}
    \caption{\textbf{Qualitative Comparison.} Comparison results between various state-of-the-art detection methods and the proposed method are shown above. Our method effectively leverages modality context to significantly enhance anomaly localization (highlighted in red), compared to baseline results (highlighted in blue). Ground truth bounding boxes are highlighted in green. For cases where the bounding boxes are small, insets show an enlarged view of the highlighted yellow region.}
    \label{fig:qualitative}
\end{figure*}

\paragraph{Qualitative Results.}
Fig.~\ref{fig:qualitative} compares our proposed framework with various detection methods across multiple modalities.
Baseline detectors frequently miss subtle lesions or produce overly large bounding boxes.
Our approach enforces modality-aware semantics and improves the precision of bounding box placement to successfully recover missed findings and produce clinically meaningful boxes.
Additional results are included in Appendix~\ref{sec:appendix-qualitative}.

\paragraph{Alternative token designs.}
We isolate what the modality tokens actually contribute through a set of token-level controls (Tab.~\ref{tab:modality-tokens}).
All variants share architecture and training schedule; construction details are in Appendix~\ref{sec:appendix-token}.
Two capacity controls, a single input-agnostic learnable token and per-modality random vectors, remain within 0.2 AP of the baseline, so neither the extra attention slot nor the added parameters accounts for our gains. Because the random vectors induce exactly the partition used by \texttt{\{MODALITY\}}, holding that partition fixed and replacing them with text-derived tokens improves AP by 2.8.
A per-dataset token, naming both dataset and modality (\textit{e.g.}, ``VinBig chest X-ray dataset'') over a partition strictly finer than modality, falls 2.7 AP short of \texttt{\{MODALITY\}}, so source and task identity do not explain the improvement.
Shuffling the tokens so that they no longer match the input modality retains only part of the benefit (38.3 AP): a consistently assigned but incorrect token still partitions the query space by modality, and correctly assigning it adds a further 2.4 AP.
Notably, \texttt{\{MODALITY\}} alone reaches 40.7 AP with no class information supplied at training or inference, so the gains stem from alignment to a semantically structured text space rather than from dataset identity or simple partition.

\paragraph{Comparison with modality-specific experts.}
A natural alternative to a unified detector is to train one expert per modality and route each image by its acquisition metadata. Tab.~\ref{tab:specialist-vs-unified} compares the two. The concern is well founded for a plain unified detector, which falls behind the expert ensemble on both backbones, confirming that the interference cost of mixing modalities is real~\cite{guan2021domain, kang2023structure, yang2024cross}. Making modality context explicit reverses this: our method exceeds the ensemble by 2.0 AP on DINO while hosting a fifth of the parameters and requiring one tuning cycle rather than five~\cite{ickler2023taming}.
The per-modality breakdown is consistent with a supervision effect. Experts retain an edge on CT and MRI, which supply the most training images per category, whereas the unified model is stronger on CXR, pathology and colonoscopy, where images per category are scarce. Sharing appears to pay where annotation is the binding constraint, the common situation in clinical practice.

\paragraph{Runtime Comparison.}
The proposed MoCA induces minimal increase in inference time as it incorporates only a single extra token for self-attention. Specifically, we compare decoder head latency (ms) for the two baseline models used (\textit{i.e.}, Deformable DETR, DINO) in Tab.~\ref{tab:decoder_latency}.
The reported values are measurements of the average time required per image, measured over 100 samples.

\begin{table*}[t]
\centering
\scriptsize
\caption{
Comparison of modality-specific expert detectors and unified multimodality detectors.
Specialist totals assume each image is sent to the expert for its own modality with no routing errors, and predictions are pooled over the full 27-category taxonomy.
\textbf{Bold} = best per column within each detector.
}
\label{tab:specialist-vs-unified}
\setlength{\tabcolsep}{3pt}
\begin{tabular}{llcccccccccccccc}
\toprule
& & \multicolumn{2}{c}{Deployment} & \multicolumn{2}{c}{Total} & \multicolumn{2}{c}{CXR} & \multicolumn{2}{c}{MRI} & \multicolumn{2}{c}{Colonoscopy} & \multicolumn{2}{c}{Pathology} & \multicolumn{2}{c}{CT} \\
\cmidrule(lr){3-4}\cmidrule(lr){5-6}\cmidrule(lr){7-8}\cmidrule(lr){9-10}\cmidrule(lr){11-12}\cmidrule(lr){13-14}\cmidrule(lr){15-16}
Detector & Method & Models & Params (M) & AP & AP$_{50}$ & AP & AP$_{50}$ & AP & AP$_{50}$ & AP & AP$_{50}$ & AP & AP$_{50}$ & AP & AP$_{50}$ \\

\midrule
\multirow{3}{*}{Def.\ DETR}
& Baseline (Specialist)
& 5 & 200.50 & 36.2  & 55.4 & 14.9 & 30.9 & \textbf{77.4} & \textbf{96.7} & \textbf{69.0} & 84.4 & \textbf{47.3} & 74.1 & 31.0 & \textbf{63.6} \\
\cmidrule(lr){2-16}
& Baseline (Unified)
& 1 & 40.11 & 35.0 & 54.4 & 15.1 & 30.9 & 75.8 & 96.2 & 61.3 & 77.3 & 46.5 & 74.1 & 30.5 & 62.3 \\
\rowcolor{LightBlue}
& + Ours
& 1 & 40.89 & \textbf{37.2} & \textbf{60.2} & \textbf{18.2} & \textbf{39.9} & 76.1 & \textbf{96.7} & 66.7 & \textbf{84.6} & 45.4 & \textbf{76.0} & \textbf{31.3} & 61.6 \\

\midrule
\multirow{3}{*}{DINO}
& Baseline (Specialist)
& 5 & 237.70 & 39.1  & 59.1  & 17.2 & 34.6 & \textbf{80.4} & 97.2 & 69.9 & 84.3 & 52.9 & 81.9 & \textbf{36.5} & \textbf{70.6} \\
\cmidrule(lr){2-16}
& Baseline (Unified)
& 1 & 47.55 & 37.7 & 58.6 & 17.8 & 36.1 & 78.0 & 96.6 & 65.1 & 81.3 & 50.5 & 77.8 & 33.0 & 66.6 \\
\rowcolor{LightBlue}
& + Ours
& 1 & 48.34& \textbf{41.1} & \textbf{65.5} & \textbf{21.1} & \textbf{45.4} & 78.6 & \textbf{97.4} & \textbf{71.2} & \textbf{88.7} & \textbf{55.5} & \textbf{85.4} & 34.4 & 67.5 \\

\bottomrule
\end{tabular}
\end{table*}

\begin{figure*}[t]
\centering
\begin{minipage}[t]{0.72\linewidth}\vspace{0pt}
    \centering
    \captionof{table}{Effectiveness of QueryREPA in utilization of unlabeled datasets. Datasets are: (1) NIH with only class labels (2) NIH with bounding box labels (3) VinBigData with bounding box labels. \best{Bold} = best per column.}
    \vspace{-0.3cm}
    \label{tab:nih}
    \begin{center}
    \footnotesize
    \setlength{\tabcolsep}{5pt}
    \renewcommand{\arraystretch}{1.2}
    \resizebox{\linewidth}{!}{%
    \begin{tabular}{cc|cc|cc|cc|cc|cc|cc}
    \toprule
    QueryREPA & Detection Training &
    \multicolumn{2}{c|}{Total} & \multicolumn{2}{c|}{CXR} & \multicolumn{2}{c|}{MRI} &
    \multicolumn{2}{c|}{Colonoscopy} & \multicolumn{2}{c|}{Pathology} & \multicolumn{2}{c}{CT} \\
    Dataset & Dataset & AP & AP$_{50}$ & AP & AP$_{50}$ & AP & AP$_{50}$ & AP & AP$_{50}$ & AP & AP$_{50}$ & AP & AP$_{50}$ \\
    \midrule

    \xmark & (2) + (3) &
    38.5 & 62.0 &
    19.1 & 41.5 &
    77.4 & 96.8 &
    68.7 & 86.8 &
    53.4 & 83.0 &
    34.2 & 66.2 \\
    
    (2) + (3) & (2) + (3) &
    39.1 & 62.9 &
    19.2 & 42.5 &
    77.6 & 96.6 &
    70.4 & 85.7 &
    \textbf{55.7} & \textbf{85.9} &
    \textbf{34.6} & 67.7 \\

    \rowcolor{LightBlue}
    (1) + (2) + (3) & (2) + (3) &
    \textbf{39.4} & \textbf{63.3} &
    \textbf{19.6} & \textbf{42.6} &
    \textbf{78.0} & \textbf{97.1} &
    \textbf{72.6} & \textbf{89.9} &
    54.7 & 84.8 &
    34.1 & \textbf{68.0} \\
    
    \bottomrule
    \end{tabular}}
    \end{center}
\end{minipage}\hfill
\begin{minipage}[t]{0.24\linewidth}\vspace{0pt}
    \captionof{table}{Comparison of decoder head latency (ms) of baseline models with and without MoCA.}
    \label{tab:decoder_latency}
    \begin{center}
    \resizebox{0.85\linewidth}{!}{
    \begin{tabular}{lc}
        \toprule
        \textbf{Model} & \textbf{Runtime (ms)} \\
        \midrule
        Def. DETR & 7.80 \\
        Def. DETR + Ours & 8.71 \\
        \midrule
        DINO & 9.37 \\
        DINO + Ours & 9.59 \\
        \bottomrule
    \end{tabular}
    }
    \end{center}
\end{minipage}
\end{figure*}

\subsection{QueryREPA Pretraining on Large-Scale Weakly Labeled Datasets}
\label{sec:nih}

QueryREPA performs representation alignment at the query level, requiring only image-level class labels and \textit{no bounding-box supervision during pretraining}. This makes it well suited to large medical datasets where detailed annotations are limited. As an example, we additionally leverage the NIH Chest X-ray dataset~\cite{Wang_2017}, which contains over 112,000 images but bounding-box annotations for fewer than 1,000 cases. In Tab.~\ref{tab:nih} we vary the datasets used for QueryREPA while fixing those used for detection training.

We observe that including images without bounding boxes during QueryREPA pretraining yields measurable gains, turning otherwise underutilized weakly labeled data into a useful resource. Notably, the gains are not confined to CXR: adding chest radiographs improves colonoscopy by 2.2 AP and MRI by 0.4, a form of cross-modality transfer unavailable to the modality-specific experts of Tab.~\ref{tab:specialist-vs-unified}. This suggests a broader avenue for weakly supervised learning in detection models \textit{directly on object queries}.

\begin{table}[t]
    \centering
    \caption{Ablation of components using DINO~\cite{zhang2022dino} as the base model and CLIP~\cite{radford2021learning} as the encoder $\gE$. Checkmarks denote modules enabled. \best{Bold} = best per column.}
    \label{tab:component-ablation}
    \resizebox{.85\linewidth}{!}{
    \begin{tabular}{cc|c|c|c|c|c|c}
    \toprule
    QueryREPA & MoCA \quad & AP & AP$_{50}$ & AP$_{75}$ & AP$_s$ & AP$_m$ & AP$_l$ \\
    \midrule
    & &
    37.7 & 58.6 & 37.8 & 19.9 & 29.8 & 41.7 \\
    \tick & &
    38.1 & 58.4 & 40.9 & 21.0 & 29.8 & 42.3 \\
    & \tick &
    40.8 & 65.1 & \best{43.4} & 25.1 & 33.5 & \best{45.2} \\
    \rowcolor{LightBlue}
    \tick & \tick &
    \best{41.1} & \best{65.5} & \best{43.4} & \best{25.4} & \best{34.6} & 44.9 \\
    \bottomrule
    \end{tabular}
    }
\end{table}

\subsection{Ablation Study}

Tab.~\ref{tab:component-ablation} provides an ablation of the individual components. Both MoCA and QueryREPA improve over the baseline on their own, and combining them performs best. Tab.~\ref{tab:alignment_layer_ablation} in Appendix~\ref{sec:appendix-decoder-layer} further shows that performance is largely unaffected by which decoder layer $l$ QueryREPA is applied to.

\vspace{-9mm}
\section{Conclusion}

In this work, we showed that aligning object-query representations mitigates the performance degradation that arises when detectors are trained on mixed multimodality datasets, as reflected in the disentanglement of the query space.
Our proposed framework introduces lightweight, text-derived modality tokens to provide explicit cues to the queries.
The Multimodality Context Attention (MoCA) mechanism integrates these tokens into the detection process, while the QueryREPA pretraining stage creates a modality-aware query set before downstream training.
Our experiments show that the combination of MoCA and QueryREPA consistently improves detection accuracy for a mixed set of medical imaging modalities.
We further show that a unified detector equipped with our method outperforms an ensemble of modality-specific experts while requiring a fraction of the parameters, indicating that mixed-modality training is preferable to partitioning once modality context is made explicit.
Our method offers a detector-agnostic, low-overhead, and practical solution, paving the way for the development of robust and generalizable medical object detection models in real-world clinical settings.

\paragraph{Limitations.}

Our evaluation is limited to public datasets, and further validation on multi-center clinical data is needed. Furthermore, QueryREPA requires a separate pretraining stage before end-to-end detection training, whereas future work could explore joint alignment and detection training. These limitations suggest important directions toward robust multimodality medical object detection.

\section*{Acknowledgments}
This work was supported by the National Research Foundation of Korea under Grant RS-2024-00336454.
This work was supported by the Korea University Grant (K2629711).

{
    \small
    \bibliographystyle{ieeenat_fullname}
    \bibliography{main}
}

\clearpage
\appendix

\section{Proofs}

\lowerbound*
\begin{proof}
Let $U^{(l)}$ and $V$ be the positive pair with joint $p(u,v)$ and marginals $p(u), p(v)$. Let $K$ i.i.d. negatives $\{V_k\}_{k=1}^K \sim p(v)$ be sampled independently of $U^{(l)}$. Define scores $s(u,v)=\text{sim}(u,v)/\tau$ and $r(u,v)=\exp(s(u,v))$. The InfoNCE objective of \eqref{eq:infonce} can be expressed as:
\begin{equation}
    \gL_\text{QRA}^{(l)} = -\mathbb{E}\left[ \log{\frac{r(U^{(l)}, V)}{r(U^{(l)}, V) + \sum_{k=1}^K{r(U^{(l)}, V_k)}}} \right].
    \label{eq:infonce2}
\end{equation}
Introduce a uniformly random index $J\in\{0, \cdots,K\}$ and construct the candidate set $\{ V_0, \cdots,V_K \}$ by placing the positive at slot $J(V_j \sim p(v|U^{(l)}))$ and drawing the remaining $K$ elements i.i.d. from $p(v)$. By Bayes' rule and exchangeability,
\begin{equation}
p(J=j \mid u, v_{0:K})
=\frac{\frac{p(v_j\mid u)}{p(v_j)}}{\sum_{m=0}^K \frac{p(v_m\mid u)}{p(v_m)}}.
\label{eq:A.1}
\end{equation}
The model's softmax over scores,
\begin{equation}
q_\theta(J=j \mid u, v_{0:K})
=\frac{r(u,v_j)}{\sum_{m=0}^K r(u,v_m)}.
\label{eq:A.2}
\end{equation}
is exactly the classifier used by InfoNCE.
The InfoNCE loss is the average cross-entropy between the true posterior \(p(\cdot\mid u,v_{0:K})\) and \(q_\theta(\cdot\mid u,v_{0:K})\):
\begin{equation}
\resizebox{\columnwidth}{!}{$\displaystyle
\begin{aligned}
\mathcal L_{\mathrm{QRA}}^{(l)}
&=\mathbb E\!\left[-\log q_\theta\!\left(J\mid U^{(l)},V_{0:K}\right)\right] \\
&=\mathbb E\!\left[H\!\left(p(\cdot\mid U^{(l)},V_{0:K})\right)
+ \mathrm{KL}\!\left(p\Vert q_\theta\right)\right]
\;\ge\; \mathbb E\!\left[H\!\left(J\mid U^{(l)},V_{0:K}\right)\right].
\end{aligned}$}
\label{eq:A.3}
\end{equation}
Generate \(Z=(J,V_{0:K})\) from \(V\) by adding i.i.d.\ negatives and randomly placing the positive. This defines a channel \(p(z\mid v)\) using only independent noise, so \(U^{(l)} \to V \to Z\) is a Markov chain. By the data processing inequality,
\begin{equation}
I\!\left(U^{(l)};V\right)\;\ge\; I\!\left(U^{(l)};Z\right)
\;\ge\; I\!\left(U^{(l)};J\mid V_{0:K}\right).
\label{eq:A.4}
\end{equation}
Since \(J\) is uniform and independent of \(V_{0:K}\) under the random placement,
\begin{equation}
\begin{aligned}
I\!\left(U^{(l)};J\mid V_{0:K}\right)
&= H(J\mid V_{0:K}) - H(J\mid U^{(l)},V_{0:K}) \\
&= \log(1+K) - \mathbb E\!\left[H\!\left(J\mid U^{(l)},V_{0:K}\right)\right].
\end{aligned}
\label{eq:A.5}
\end{equation}
Combining \eqref{eq:A.4} and \eqref{eq:A.5},
\begin{equation}
I\!\left(U^{(l)};V\right)
\;\ge\; \log(1+K) - \mathbb E\!\left[H\!\left(J\mid U^{(l)},V_{0:K}\right)\right].
\label{eq:A.6}
\end{equation}
Applying \eqref{eq:A.3} to \eqref{eq:A.6} yields
\begin{equation}
I\!\left(U^{(l)};V\right)
\;\ge\; \log(1+K) - \mathcal L_{\mathrm{QRA}}^{(l)},
\end{equation}
which is the desired standard lower bound.

\end{proof}

\section{Dataset Explanation}
\label{sec:appendix-dataset}
In cases where datasets provided only segmentation masks without explicit bounding boxes, we generated detection annotations by extracting bounding boxes directly from the segmentation masks. Detailed descriptions of each dataset are provided below.

\begin{figure*}[t]
\centering

\begin{minipage}{0.65\textwidth}
\centering
\captionof{table}{Dataset summary with image and annotation counts.}
\label{tab:dataset_summary}
\setlength{\tabcolsep}{4pt}
\renewcommand{\arraystretch}{1.1}
\resizebox{\textwidth}{!}{%
\begin{tabular}{l l c c c c c}
    \toprule
    \textbf{Dataset} & \textbf{Modality} & \textbf{Images} & \textbf{Resolutions} & 
    \makecell{\textbf{Train}\\\textbf{(Annotations)}} & 
    \makecell{\textbf{Test}\\\textbf{(Annotations)}} & 
    \textbf{Class} \\
    \midrule
    Vinbig & CXR & 4394 & Variable & 3296 (18024) & 1098 (5931) & 14 \\
    COVID-19 CT & CT & 1388 & 1024 × 1024 & 1187 (3801) & 201 (240) & 1 \\
    LIDC-IDRI & CT & 9122 & 1024 × 1024 & 7389 (7927) & 1733 (1957) & 1 \\
    NeoPolyp & Colonoscopy & 1000 & 1024 × 1024 & 800 (1891) & 200 (454) & 3 \\
    Br35H & MRI & 701 & Variable & 500 (564) & 201 (240) & 1 \\
    ACDC & MRI & 2827 & 1024 × 1024 & 1831 (4934) & 996 (2732) & 3 \\
    MonuSeg & Pathology & 1414 & 320 × 320 & 1017 (31367) & 397 (13749) & 4 \\
    \bottomrule
\end{tabular}}
\end{minipage}

\vspace{1.2em}

\begin{minipage}{\textwidth}
\centering
{\ttfamily
\renewcommand{\arraystretch}{1.45}
\small
\resizebox{\textwidth}{!}{%
\begin{tabular}{lll}
\textbullet\ "Aortic enlargement in CXR"  & \textbullet\ "Atelectasis in CXR"                   & \textbullet\ "Calcification in CXR" \\
\textbullet\ "Cardiomegaly in CXR"        & \textbullet\ "Consolidation in CXR"                 & \textbullet\ "ILD in CXR" \\
\textbullet\ "Infiltration in CXR"        & \textbullet\ "Lung Opacity in CXR"                  & \textbullet\ "Nodule/Mass in CXR" \\
\textbullet\ "Other lesion in CXR"        & \textbullet\ "Pleural effusion in CXR"              & \textbullet\ "Pleural thickening in CXR" \\
\textbullet\ "Pneumothorax in CXR"        & \textbullet\ "Pulmonary fibrosis in CXR"            & \textbullet\ "Brain tumor in MRI" \\
\textbullet\ "Epithelial in Pathology (H\&E stain)" & \textbullet\ "Lymphocyte in Pathology (H\&E stain)" & \textbullet\ "Neutrophil in Pathology (H\&E stain)" \\
\textbullet\ "Macrophage in Pathology (H\&E stain)" & \textbullet\ "Left heart ventricle in cardiac MRI" & \textbullet\ "Myocardium in cardiac MRI" \\
\textbullet\ "Right heart ventricle in cardiac MRI" & \textbullet\ "COVID-19 infection in lung CT"       & \textbullet\ "Nodule in lung CT" \\
\textbullet\ "Neoplastic polyp in colon endoscope"  & \textbullet\ "Polyp in colon endoscope"            & \textbullet\ "Non-neoplastic polyp in colon endoscope" \\
\end{tabular}}}
\caption{List of 27 categories used in our experiments.}
\label{fig:class_list_vertical}
\end{minipage}

\end{figure*}

\setlength{\tabcolsep}{4pt} 
\renewcommand{\arraystretch}{1.1} 

\paragraph{VinBigData Chest Abnormalities Dataset (CXR).}
The VinBigData Chest Abnormalities Dataset~\cite{vinbig} contains chest X-ray images annotated by multiple radiologists, aimed at developing automated detection systems for chest abnormalities. The dataset has been converted to COCO format and optimized using the Weighted Boxes Fusion (WBF) technique. It includes 14 classes of abnormalities, such as aortic enlargement, cardiomegaly, lung opacity, and pneumothorax. Images in the dataset have varying resolutions and are divided into training and testing collections.

\paragraph{COVID-19 CT Dataset (CT).}
The COVID-19 CT dataset~\cite{covid19ct} consists of CT scans containing COVID-19 infections with pixel-level annotations. This dataset is used for segmentation and classification of COVID-19 related lung abnormalities. All images have a resolution of 1024 × 1024 pixels and are split into training and testing sets.

\paragraph{LIDC-IDRI Dataset (CT).}
The LIDC-IDRI dataset~\cite{lidc} is a large-scale collection of lung CT scans with nodule annotations from four experienced radiologists. It is widely used in lung nodule detection and segmentation research. The dataset follows the COCO format with images of 1024 × 1024 resolution for training and testing purposes.

\paragraph{NeoPolyp Dataset (Colonoscopy).}
The NeoPolyp dataset~\cite{neopolyp} contains colonoscopy images annotated for polyp detection and classification. This dataset provides images of resolution 1024 × 1024 and supports training and testing sets, facilitating the development of deep learning models for polyp detection.

\paragraph{BR35H Brain Tumor Dataset (MRI).}
The BR35H dataset~\cite{br35h} consists of brain MRI scans categorized into tumor and non-tumor cases. The dataset is used for brain tumor detection tasks. The images have variable resolutions and are split into training and testing sets.

\paragraph{ACDC Dataset (MRI).}
The ACDC dataset~\cite{acdc} provides cardiac MRI images with expert annotations for segmentation and classification of heart structures. It consists of 1024 × 1024 resolution images and supports both training and testing configurations.

\paragraph{MoNuSeg Dataset (Pathology).}
The MoNuSeg dataset~\cite{monuseg} is designed for nuclear segmentation in histopathology images. It includes diverse samples from multiple organs, enabling the development of robust segmentation models. Images are provided at a resolution of 320 × 320 pixels and are divided into training and testing sets.

\section{Training Details}
\label{sec:training-details}
\paragraph{Setup for Deformable DETR with MoCA and QueryREPA.}
We train ResNet-50-based Deformable DETR for 50 epochs using AdamW (lr=$1{\times}10^{-4}$, wd=$1{\times}10^{-4}$) with a MultiStep learning-rate decay at epoch~40. 
We use 300 object queries throughout training. 
Pretraining optimizes only a modality-balanced contrastive loss 
(QueryREPA), sampling mini-batches with a \texttt{ModalityBatchSampler} to ensure balanced modality coverage. 
Data augmentation follows the standard DETR recipe, 
including random horizontal flip, multi-scale random resize 
of the shorter image side within [480, 800] px, and 
random absolute-range cropping with re-resizing.
Finetuning resumes from the final pretraining checkpoint, 
training with Hungarian matching using focal loss (weight=2.0, $\alpha{=}0.25,\gamma{=}2.0$), 
L1 loss (weight=5.0), and GIoU loss (weight=2.0). 
We use a per-GPU batch size of 4 with the default sampler.
For positional encoding in self-attention layers, we apply zero padding to maintain compatibility with architecture.

\paragraph{Setup for DINO with MoCA and QueryREPA.}
We train ResNet-50-based DINO for 36 epochs using AdamW (lr=$2{\times}10^{-4}$, wd=$1{\times}10^{-4}$) with a MultiStep learning-rate decay at epoch~30. 
Each iteration uses 900 object queries and 100 denoising queries. 
Pretraining optimizes only a modality-balanced contrastive loss 
(QueryREPA) with a \texttt{ModalityBatchSampler} for balanced sampling. 
The same data augmentation pipeline as above is applied. 
Finetuning resumes from the final pretraining checkpoint, 
training with Hungarian matching using focal loss (weight=2.0, $\alpha{=}0.25,\gamma{=}2.0$), 
L1 loss (weight=5.0), and GIoU loss (weight=2.0). 
We use a per-GPU batch size of 4 with the default sampler.
For positional encoding in self-attention layers, we apply zero padding to maintain compatibility with architecture.

\paragraph{Text encoder and embedding generation.}
Each image is represented by a \texttt{\{CLASS in MODALITY\}} prompt 
(see Figure~\ref{fig:class_list_vertical}), with categories deduplicated and sorted. 
We precompute embeddings with three frozen encoders 
(OpenCLIP ViT-B/32, BiomedCLIP, PubMedCLIP) by tokenizing and encoding the 
prompts in \texttt{eval()} mode on GPU with \texttt{torch.no\_grad()}, 
saving one \texttt{.npy} per image. 
During finetuning, these embeddings are loaded per image and used to compute 
the contrastive alignment loss.

\paragraph{Choice of Text.}
We adopt a concise and unambiguous textual representation for each category.
Specifically, we employ the \texttt{\{CLASS in MODALITY\}} template, which pairs
the semantic name of the target class (\texttt{CLASS}) with its imaging domain
(\texttt{MODALITY}), or the \texttt{\{MODALITY\}} template. The complete set of 27 categories for \texttt{\{CLASS in MODALITY\}} is provided in
Figure~\ref{fig:class_list_vertical}. This formulation explicitly preserves both
object semantics and modality context, without relying on additional descriptive
phrases.

\section{Alternative Token Designs}
\label{sec:appendix-token}

Tab.~\ref{tab:modality-tokens} of the main paper establishes that the gains from our modality tokens cannot be explained by architectural capacity, correctness of assignment, or dataset-level identity, through four token-level controls (Learnable, Random, Shuffled, Per-dataset) compared against \texttt{\{MODALITY\}}. Here we provide the construction details for each variant, together with an extended comparison against \texttt{\{CLASS\}} and \texttt{\{CLASS in MODALITY\}} that further characterizes the semantic granularity required for these gains.

\paragraph{Learnable.} A learnable token is introduced in place of the text embedding: one vector per decoder layer, randomly initialized ($\mathcal{N}(0, 0.02^2)$) and shared across all images regardless of modality. It is prepended to the query sequence in self-attention and discarded before the residual connection (Eq.~6), so it modulates the object queries without occupying an output slot. This variant follows the same two-stage protocol as our full model, with the learnable token, rather than a text-derived embedding, serving as the MoCA prompt at both pretraining and fine-tuning; the CLIP text encoder is not used to construct this prompt, though it continues to supply the contrastive target during QueryREPA pretraining as in our main design.

\paragraph{Random.} A fixed, non-learnable random vector is sampled independently for each imaging modality and held constant throughout training and inference (\textit{i.e.}, the same vector is reused for every image of a given modality), whose L2 norm is set to the mean norm of the corresponding real text-derived embeddings, so that any performance difference reflects semantic content rather than a mismatch in vector scale. The prompt token enters MoCA's self-attention without normalization, making the attention logits sensitive to its magnitude.

\paragraph{Shuffled.} We use the same, genuine CLIP-derived embeddings for each modality as in our main design, but permute the assignment between modalities and embeddings using a derangement (no modality is mapped to itself), applied identically across training and inference. Every image therefore receives a syntactically valid but factually incorrect modality embedding. This is the same manipulation used to construct the swapped-token condition in the attention analysis of Fig.~\ref{fig:attention_heatmap}, here applied at the level of training and evaluation rather than a single qualitative example.

\paragraph{Per-dataset.} We construct a text prompt naming both the source dataset and the modality (\textit{e.g.}, ``VinBig chest X-ray dataset''), encoded with the same CLIP text encoder as our main tokens. Because several modalities in our benchmark aggregate multiple source datasets, this induces a partition strictly finer than modality; the resulting tokens are still text-derived and carry genuine semantic content, but at the granularity of dataset identity rather than imaging modality.

\paragraph{Extended comparison with class-level tokens.} Tab.~\ref{tab:modality-tokens} additionally reports \texttt{\{CLASS\}} and \texttt{\{CLASS in MODALITY\}}, following the same text-derived construction as \texttt{\{MODALITY\}} but using the target class alone, or the class and modality jointly, as described in Sec.~3.1. Jointly encoding class and modality yields the best result overall, indicating that the tokens capture fine-grained, class-specific distinctions beyond a coarse modality label.

\section{Ablation of Decoder Layer $l$}
\label{sec:appendix-decoder-layer}
We further investigate the effect of applying QueryREPA on different decoder layers of the detection head. As shown in Tab.~\ref{tab:alignment_layer_ablation}, our method is capable of producing optimal results regardless of decoder layer $l$. In our main experiments, we choose layer index $l=5$, due to its slight increase ($+0.1$) in total AP.

\begin{table}[t]
    \centering
    \caption{Ablation of decoder layer $l$ used for alignment in QueryREPA. Our method is robust, producing optimal results regardless of decoder layer $l$.}
    \label{tab:alignment_layer_ablation}
    \resizebox{0.85\linewidth}{!}{
    \begin{tabular}{l | cccccc}
    \toprule
    \textbf{Layer $l$}\quad & \ \textbf{AP} & \textbf{AP$_{50}$} & \textbf{AP$_{75}$} & \textbf{AP$_s$} & \textbf{AP$_m$} & \textbf{AP$_l$} \\
    \midrule
    Layer 2 \quad & \ 41.0 & 65.8 & 43.5 & 24.0 & 33.3 & 45.0 \\
    Layer 3 \quad & \ 41.0 & 65.6 & 43.3 & 27.0 & 32.8 & 45.1 \\
    Layer 4 \quad & \ 41.0 & 65.8 & 43.1 & 25.3 & 33.3 & 45.3 \\
    Layer 5 \quad & \ 41.1 & 65.5 & 43.4 & 25.4 & 34.6 & 44.9 \\
    \bottomrule
    \end{tabular}
    }
\end{table}

\section{Beyond DETR-style Models}
Our framework is highly general and can be applied whenever two mild conditions are met: (i) object-level embeddings are used, and (ii) these embeddings undergo self-attention. These requirements are very loose and are satisfied by most modern detectors, including CNN-style and diffusion-based methods.

For example, traditional CNN-style detectors (\textit{e.g.}, Faster R-CNN~\cite{ren2015faster}, Mask R-CNN~\cite{He_2017_ICCV}) originally relied on a large number of dense object candidates, which led to high training cost. More recent CNN-style detectors (\textit{e.g.}, Sparse R-CNN~\cite{sun2021sparse}, DiffusionDet~\cite{chen2023diffusiondet}) alleviate this inefficiency and improve performance by introducing a small set of object proposals with attention-based iterative refinement. These object proposals are analogous to the object-level embeddings (\textit{i.e.}, ``object queries'' in DETR-style methods) that we exploit in our work. Hence, our method can also be readily adapted for modern CNN-style detectors.

As a proof of concept, we show additional results applying our method (QueryREPA\allowbreak{+MoCA)} to the state-of-the-art detector DiffusionDet in Tab.~\ref{tab:diffusiondet}. Note that DiffusionDet directly adopts the detection decoder from Sparse R-CNN and can therefore be viewed as a CNN-style detector. To integrate our method into DiffusionDet, we simply change its region proposals (akin to object proposals) into learnable embeddings and then apply our approach: we concatenate a modality token in the self-attention stage (MoCA) and perform QueryREPA pre-training. This lightweight adaptation substantially improves detection performance, for example boosting AP$_{50}$ from 60.3 to 65.4 (+5.1). These results demonstrate that our method can be easily incorporated into most modern detectors while still yielding significant performance gains.

\begin{table}[t]
    \caption{Quantitative comparison of applying \colorbox{LightBlue}{our method} (MoCA + QueryREPA) on the DiffusionDet~\cite{chen2023diffusiondet} model. \best{Bold} = best per column. }
    \label{tab:diffusiondet}
    \begin{center}
    \resizebox{1.0\linewidth}{!}{
    \begin{tabular}{l|c|c|c|c|c|c}
    \toprule
    Method & AP & AP$_{50}$ & AP$_{75}$ & AP$_s$ & AP$_m$ & AP$_l$ \\
    \midrule
    DiffusionDet &
    38.2 & 60.3 & 40.1 & 22.4 & 30.6 & 41.6 \\
    \rowcolor{LightBlue}
    DiffusionDet + Ours ($\gE$: CLIP) &
    \textbf{40.2} & \textbf{65.4} & \best{41.6} & \best{26.0} & \best{34.5} & \best{42.6} \\
    
    \bottomrule
    \end{tabular}
    }
    \end{center}
\end{table}

\section{Zero-Shot Generalization to an External Dataset}
\label{sec:appendix-external}
To probe generalization beyond our training distribution, we evaluate our model on an \emph{external} chest X-ray detection dataset in a strict zero-shot setting.

\paragraph{Dataset.}
We use ChestX-Det~\cite{liu2020chestxdet10}, a subset of NIH ChestX-ray14~\cite{Wang_2017} in which three board-certified radiologists provide bounding-box annotations for 13 thoracic abnormality classes. Its test split contains 553 images with 1{,}734 boxes; we restrict evaluation to the 10 classes that overlap with our CXR taxonomy, keeping the 480 test images with at least one annotation in these classes (1{,}490 boxes in total). No ChestX-Det image is seen during training; the model is transferred directly from our mixed multimodality corpus without any fine-tuning.

\paragraph{Results.}
As reported in Tab.~\ref{tab:external}, both components remain effective under this out-of-distribution shift. MoCA alone raises AP from $3.6$ to $7.4$ and AP$_{50}$ from $10.0$ to $20.7$, and combining it with QueryREPA yields the best transfer ($8.0$ AP, $22.2$ AP$_{50}$), about a $+122\%$ relative improvement on both metrics. The absolute scores are low, as expected for zero-shot transfer to a dataset with different annotation protocols and label definitions, but the consistent relative gains indicate that the modality-aware representations learned by MoCA and QueryREPA transfer beyond the training distribution.

\begin{table}[t]
    \centering
    \footnotesize
    \caption{Zero-shot transfer on ChestX-Det. Models trained on our mixed corpus are evaluated without any fine-tuning. \best{Bold} = best per column.}
    \label{tab:external}
    \begin{tabular}{cc|cc}
        \toprule
        QueryREPA & MoCA & AP & AP$_{50}$ \\
        \midrule
              &       & 3.6 & 10.0 \\
        \tick &       & 4.1 & 11.3 \\
              & \tick & 7.4 & 20.7 \\
        \rowcolor{LightBlue}
        \tick & \tick & \best{8.0} & \best{22.2} \\
        \bottomrule
    \end{tabular}
\end{table}

\section{Addressing the Heterogeneity of Representation Spaces}
\label{sec:appendix-hetero}

The “raw” query manifold is originally learned purely from heterogeneous image statistics.
Our method is explicitly designed to address the heterogeneity of representation spaces across imaging modalities at the level of \textit{object queries} in modern detectors.
In our work, MoCA injects modality context into the query set by appending a text-derived modality token and mixing it with object queries. This converts the raw query manifold into a space where each query is conditioned on modality-aware semantic anchors. QueryREPA is further proposed as a way to reshape this manifold by aligning the mean query representation of each sample to its corresponding modality token, maximizing mutual information between queries and modality tokens. This effectively provides the detection decoder with more informative query initialization for downstream detection.

\paragraph{Object Query Visualization Before/After QueryREPA}
QueryREPA guides the object queries toward a modality-aware semantic representation. As shown in Figure~\ref{fig:umap-queryrepa}, the queries before QueryREPA are highly entangled and exhibit substantial overlap across modalities, reflecting the absence of modality-specific organization in the initial query space. After applying QueryREPA, the queries are appropriately mapped into disentangled regions. Such improvement is also evident quantitatively, with the silhouette score increasing from 0.489 to 0.747. These results indicate that QueryREPA reshapes the query space to incorporate modality-aware semantics, which in turn provides the decoder with more informative query initialization and facilitates more reliable reasoning across heterogeneous medical domains.

\section{Attention Heatmap Visualization}
\label{sec:appendix-attention-heatmap}
We visualize where the object queries attend in the deformable decoder of DINO~\cite{zhang2022dino}. We extract attention at decoder layer~5, where each query samples 128 locations ($8$ heads $\times\,4$ levels $\times\,4$ points) with associated weights, splat these onto a 2D grid via bilinear splatting, weight each query by its predicted-class confidence to suppress background queries, and aggregate over all 900 queries before upsampling to the input resolution.

\begin{figure*}[t]
    \centering
    \includegraphics[width=0.9\linewidth]{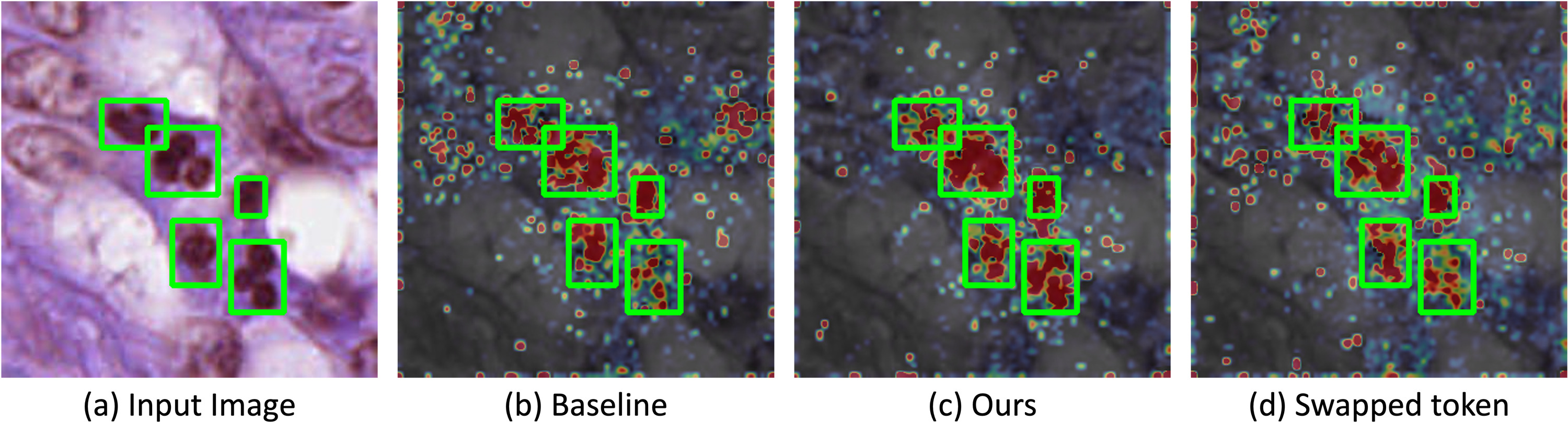}
    \caption{Attention heatmaps at decoder layer~5 (each query weighted by predicted-class confidence). (a) Input Image with ground-truth boxes; (b) Baseline; (c) Ours with the correct modality token, where attention concentrates on the target structures; (d) Swapped token, an unrelated token (\textit{``COVID-19 in lung CT''}), where the attention scatters. This indicates that the modality token causally shifts the attention pattern.}
    \label{fig:attention_heatmap}
\end{figure*}

As shown in Figure~\ref{fig:attention_heatmap}, attaching the correct modality token (c; Ours) concentrates attention more tightly on the target structures than the baseline (b). To verify that this effect is caused by the modality token rather than the image alone, we keep the input fixed and swap in a token from an unrelated modality (d; \textit{``COVID-19 in lung CT''} for a pathology image). The attention then scatters, giving causal evidence that modality tokens actively steer the queries' spatial attention toward modality-specific features.

\begin{figure}[t]
    \centering
    \includegraphics[width=\linewidth]{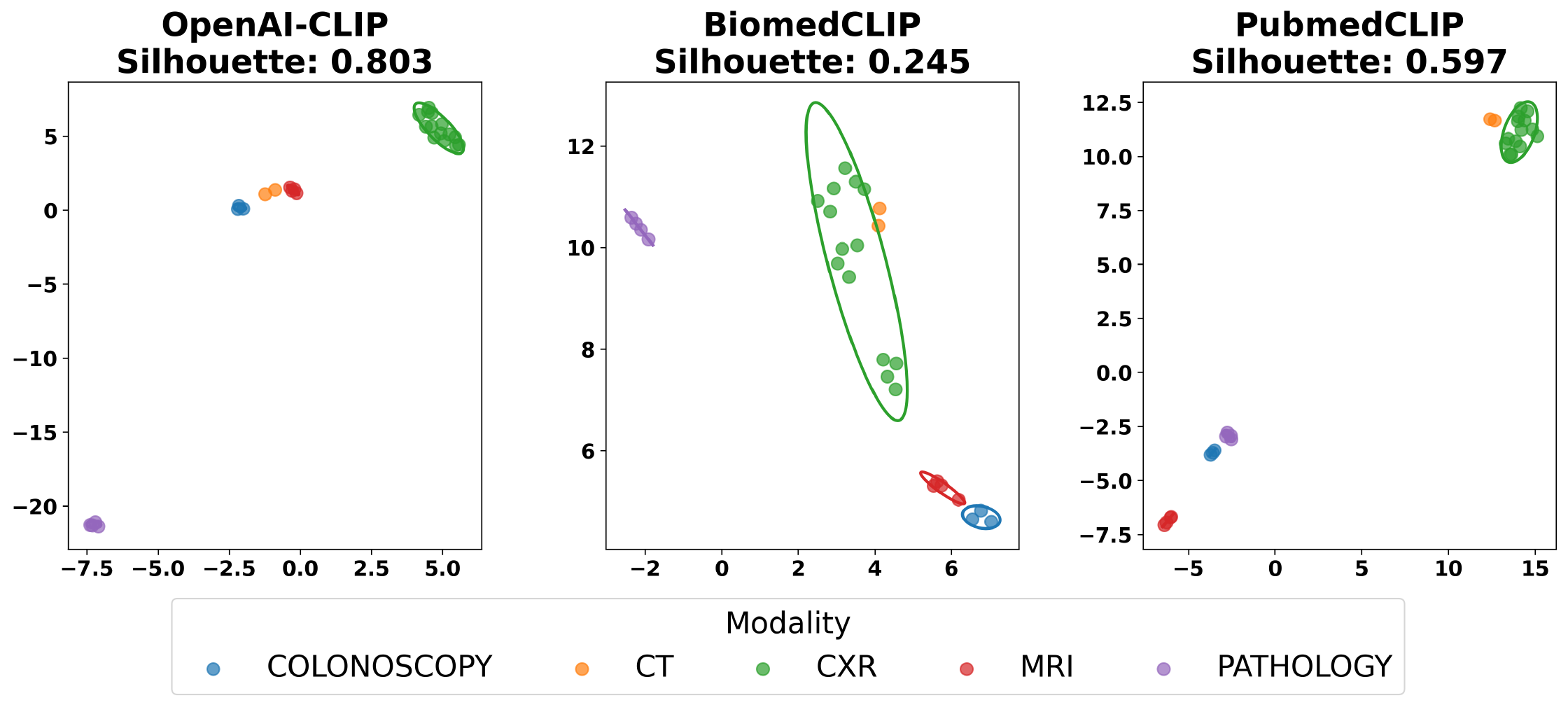}
    \caption{UMAP embedding of modality tokens for varying $\gE$.}
    \label{fig:umap}
\end{figure}

\section{Different Modality Text Encoders}
\label{sec:appendix-encoders}
\paragraph{Comparison of using different $\gM_\gE$.}
The robustness of our method allows us to use varying sets of modality tokens $\gM_\gE$ by differing the type of encoder $\gE$ used. Our quantitative comparisons in Tab.~\ref{tab:1} contain results for varying $\gM_\gE$ by selecting $\gE$ as either CLIP, BiomedCLIP, or PubMedCLIP. Though our framework provides consistent performance increase regardless of $\gE$, we find CLIP and PubMedCLIP to consistently perform better than BiomedCLIP.
We attribute this to the better disentanglement of modality tokens in CLIP and PubMedCLIP as compared to BiomedCLIP (visualized in Fig.~\ref{fig:umap}). Their corresponding silhouette scores support that better disentanglement between modalities leads to improved performance.

\section{Algorithm for QueryREPA}
\label{sec:appendix-pseudocode}
\lstset{
  columns=fullflexible,
  keepspaces=true,
  upquote=true,
  basicstyle=\ttfamily\scriptsize,
  breaklines=true,
  showstringspaces=false,
  tabsize=4,
  morecomment=[s]{"""}{"""},
  commentstyle=\color{green!60!black}
}

\begin{algorithm}[]
\scriptsize
\caption{Pseudocode of QueryREPA pretraining}
\label{code:queryrepa}
\definecolor{codered}{rgb}{0.8,0.1,0.1}
\begin{lstlisting}[language=Python, stringstyle=\color{green!60!black}]
"""
<Input>
images      : (B,H,W,3)
text_prompts: B text prompts of the form 
              '{CLASS} in {MODALITY}'
text_encoder: CLIP / BiomedCLIP / PubMedCLIP

<Output>
loss_qrepa: query representation alignment loss
# Dt: text-embedding dim from the text encoder
# D : projected query/text dim for alignment
# Nq: number of learnable object queries per image
"""

def queryrepa_loss(images, text_prompts):

    # 1) Encode text prompts into modality tokens m
    m = text_encoder(text_prompts)# (B, Dt)
    m_proj = proj_text(m)         # (B, D)
    
    # 2) Forward backbone and decoder up to layer l
    feats = detector_backbone(images)
    learnable_object_queries = init_queries(B, Nq)
                                  # (B, Nq, D)
    queries_list = detector_decoder(
        learnable_object_queries, m_proj, feats)
    q_l = queries_list[l]         # queries at layer l
    
    # 3) Mean over all queries
    q_mean = q_l.mean(dim=1)      # (B, D)
    
    # 4) Project queries into modality-token space
    q_proj = proj_query(q_mean)   # (B, Dt)
    
    # 5) InfoNCE-style alignment between q_proj and m
    q_norm = F.normalize(q_proj, dim=-1)
    m_norm = F.normalize(m, dim=-1)
    logits  = torch.matmul(q_norm, m_norm.T) / tau
    targets = torch.arange(B)
    loss_qrepa = F.cross_entropy(logits, targets)
    
    return loss_qrepa
\end{lstlisting}
\end{algorithm}

\begin{table*}[t]
\caption{Quantitative comparison leveraging various modality token encoders $\gE$ as CLIP~\cite{radford2021learning}, BiomedCLIP~\cite{zhang2023biomedclip}, and PubMedCLIP~\cite{eslami2023pubmedclip} under the same training setup. \colorbox{LightBlue}{Our method} shows improved results regardless of $\gE$ used, proving its robustness and wide applicability. \best{Bold} indicates best results per column.}
\label{tab:1}
\begin{center}
\setlength{\tabcolsep}{5pt}
\renewcommand{\arraystretch}{1.2}
\resizebox{\linewidth}{!}{%
\begin{tabular}{lcc|cc|cc|cc|cc|cc|cc}
\toprule
\multicolumn{3}{c|}{} &
\multicolumn{2}{c|}{Total} & \multicolumn{2}{c|}{CXR} & \multicolumn{2}{c|}{MRI} &
\multicolumn{2}{c|}{Colonoscopy} & \multicolumn{2}{c|}{Pathology} & \multicolumn{2}{c}{CT} \\
Method & $\gE$ & $\gL_\text{QRA}$& AP & AP$_{50}$ & AP & AP$_{50}$ & AP & AP$_{50}$ & AP & AP$_{50}$ & AP & AP$_{50}$ & AP & AP$_{50}$ \\
\midrule

\multirow{7}{*}{Deformable DETR}
& \xmark & \xmark &
35.0 & 54.4 &
15.1 & 30.9 &
75.8 & 95.9 &
61.3 & 77.3 &
46.5 & 74.1 &
30.5 & 62.3 \\

& \multirow{2}{*}{CLIP} & \xmark &
\best{37.3} & \best{60.2} &
18.0 & 39.1 &
75.3 & 96.2 &
64.6 & 83.7 &
\best{49.3} & \best{79.1} &
\best{31.5} & 62.5 \\

& & \cg \tick &
\cg 37.2 & \cg \best{60.2} &
\cg \best{18.2} & \cg \best{39.9} &
\cg \best{76.1} & \cg \best{96.7} &
\cg \best{66.7} & \cg 84.6 &
\cg 45.4 & \cg 76.0 &
\cg 31.3 & \cg 61.6 \\

& \multirow{2}{*}{BiomedCLIP} & \xmark &
36.7 & \best{60.2} &
17.5 & 39.1 &
74.5 & \best{96.7} &
65.1 & \best{85.4} &
47.7 & 77.6 &
29.9 & 61.8 \\

& & \cg \tick &
\cg 36.7 & \cg 59.6 &
\cg 17.7 & \cg 38.9 &
\cg 75.7 & \cg 96.5 &
\cg 64.2 & \cg 83.4 &
\cg 47.1 & \cg 76.8 &
\cg 30.1 & \cg 61.0 \\

& \multirow{2}{*}{PubMedCLIP} & \xmark &
36.6 & 59.7 &
17.8 & 39.0 &
75.5 & 96.4 &
64.0 & 84.4 &
46.3 & 76.1 &
29.8 & 61.0 \\

& & \cg \tick &
\cg 36.8 & \cg 60.1 &
\cg 17.8 & \cg 39.5 &
\cg 75.2 & \cg 96.4 &
\cg 64.5 & \cg 82.9 &
\cg 47.1 & \cg 77.1 &
\cg 30.7 & \cg \best{63.4} \\
\midrule

\multirow{7}{*}{DINO}
& \xmark & \xmark &
37.7 & 58.6 &
17.8 & 36.1 &
78.0 & 96.6 &
65.1 & 81.3 &
50.5 & 77.8 &
33.0 & 66.6 \\

& \multirow{2}{*}{CLIP} & \xmark &
40.8 & 65.1 &
21.1 & 45.3 &
77.0 & 96.9 &
70.4 & 87.2 &
55.5 & 85.1 &
\best{35.1} & 67.2 \\

& & \cg \tick &
\cg 41.1 & \cg 65.5 &
\cg 21.1 & \cg 45.4 &
\cg \best{78.6} & \cg \best{97.4} &
\cg 71.2 & \cg 88.7 &
\cg \best{55.5} & \cg \best{85.4} &
\cg 34.4 & \cg \best{67.5} \\

& \multirow{2}{*}{BiomedCLIP} & \xmark &
40.9 & \best{65.7} &
21.2 & 46.3 &
77.9 & 97.2 &
71.3 & 88.0 &
55.1 & 84.7 &
34.2 & 66.9 \\

& & \cg \tick &
\cg 40.6 & \cg 65.4 &
\cg 21.1 & \cg 45.9 &
\cg 77.6 & \cg 97.2 &
\cg 70.1 & \cg 85.9 &
\cg 54.7 & \cg 85.2 &
\cg 33.9 & \cg 67.4 \\

& \multirow{2}{*}{PubMedCLIP} & \xmark &
41.1 & 65.5 &
21.4 & 46.1 &
78.4 & 97.0 &
\best{72.7} & \best{89.2} &
53.6 & 83.4 &
35.4 & 67.3 \\

&  & \cg \tick &
\cg \best{41.3} & \cg65.5 &
\cg \best{22.0} & \cg \best{46.5} &
\cg 78.1 & \cg 97.0 &
\cg 72.4 & \cg 86.9 &
\cg 54.1 & \cg 84.0 &
\cg 33.7 & \cg 66.6 \\

\bottomrule
\end{tabular}}
\end{center}
\end{table*}

\begin{table}[!t]
\centering
\caption{Training cost on a single RTX 4090.}
\label{tab:cost}
\setlength{\tabcolsep}{4pt}
\begin{tabular}{lcc}
\toprule
Stage & Time & Peak Memory \\
\midrule
Baseline training        & $\sim$25h & 14.8 GB \\
QueryREPA training & $\sim$29h & 8.1 GB \\
Detection training (w/ MoCA) & $\sim$27h & 14.8 GB \\
\bottomrule
\end{tabular}
\end{table}

\section{Training Cost Comparison}
\label{sec:appendix-training-cost}
We quantify the additional training cost introduced by QueryREPA on a single RTX 4090 (Tab.~\ref{tab:cost}). QueryREPA pretraining takes approximately 29 hours, comparable to the 25 hours required for baseline detection training, while using substantially less peak memory (8.1 GB vs.\ 14.8 GB), owing in part to a smaller batch size. The subsequent detection fine-tuning stage with MoCA takes approximately 27 hours, essentially matching the baseline, so the one-time pretraining stage accounts for the added cost.

\newpage
\section{More Qualitative Results}
\label{sec:appendix-qualitative}

\begin{figure*}[t]
    \centering
    \includegraphics[width=.9\linewidth]{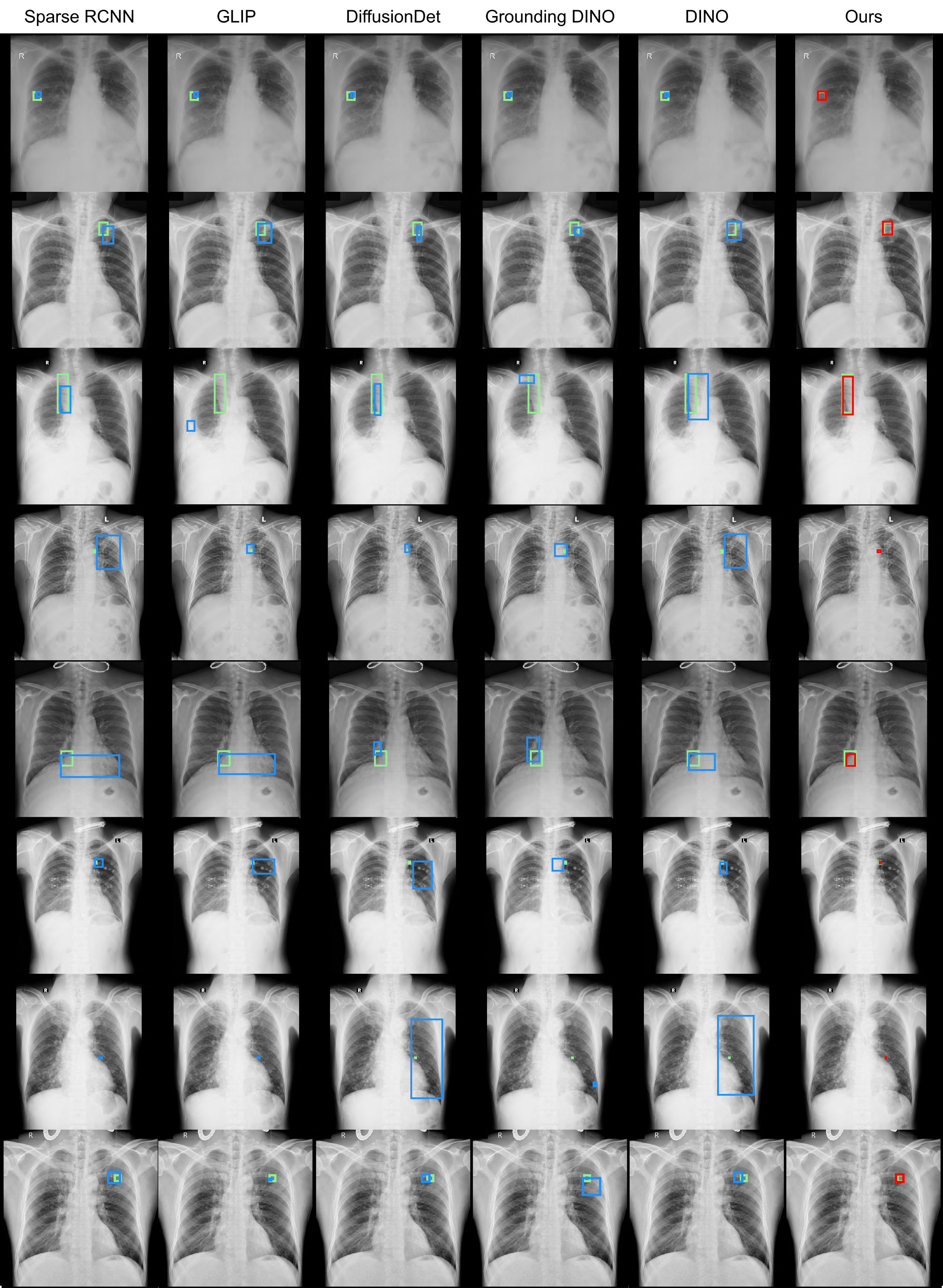}
    \caption{
        \textbf{Qualitative comparisons on \uline{chest X-ray} images.}
        \textcolor{blue}{Blue} boxes indicate predictions from baseline detectors 
        (Sparse R-CNN, GLIP, DiffusionDet, Grounding DINO, DINO), 
        \textcolor{green!60!black}{Green} boxes denote ground-truth annotations, 
        and \textcolor{red}{Red} boxes show predictions from our method.
    }
    \label{fig:cxr_qualitative}
\end{figure*}

\begin{figure*}[t]
    \centering
    \includegraphics[width=.9\linewidth]{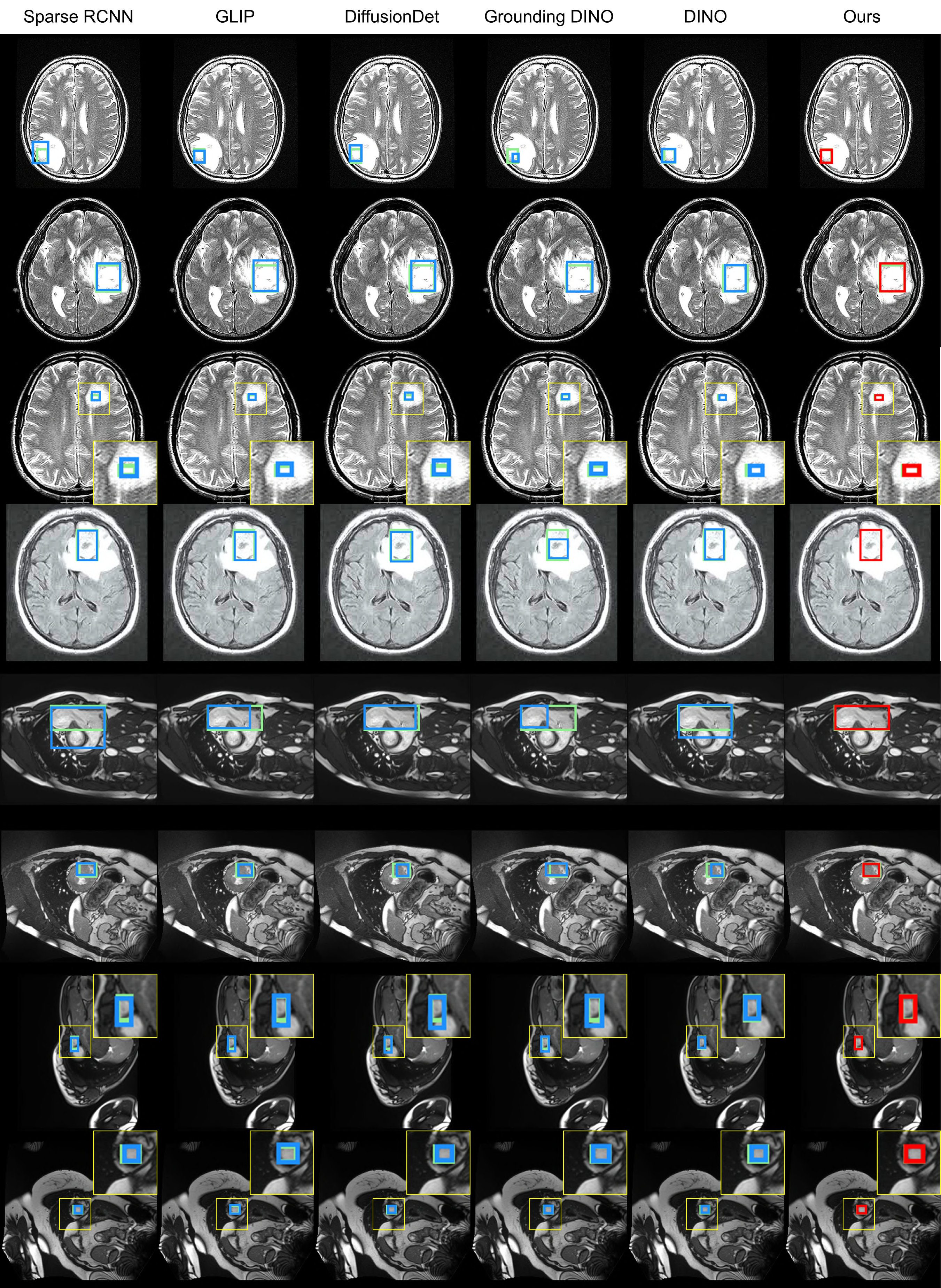}
     \caption{
        \textbf{Qualitative comparisons on \uline{MRI} images.}
        \textcolor{blue}{Blue} boxes indicate predictions from baseline detectors 
        (Sparse R-CNN, GLIP, DiffusionDet, Grounding DINO, DINO), 
        \textcolor{green!60!black}{Green} boxes denote ground-truth annotations, 
        and \textcolor{red}{Red} boxes show predictions from our method.
    }
    \label{fig:mri_qualitative}
\end{figure*}

\begin{figure*}[t]
    \centering
    \includegraphics[width=.9\linewidth]{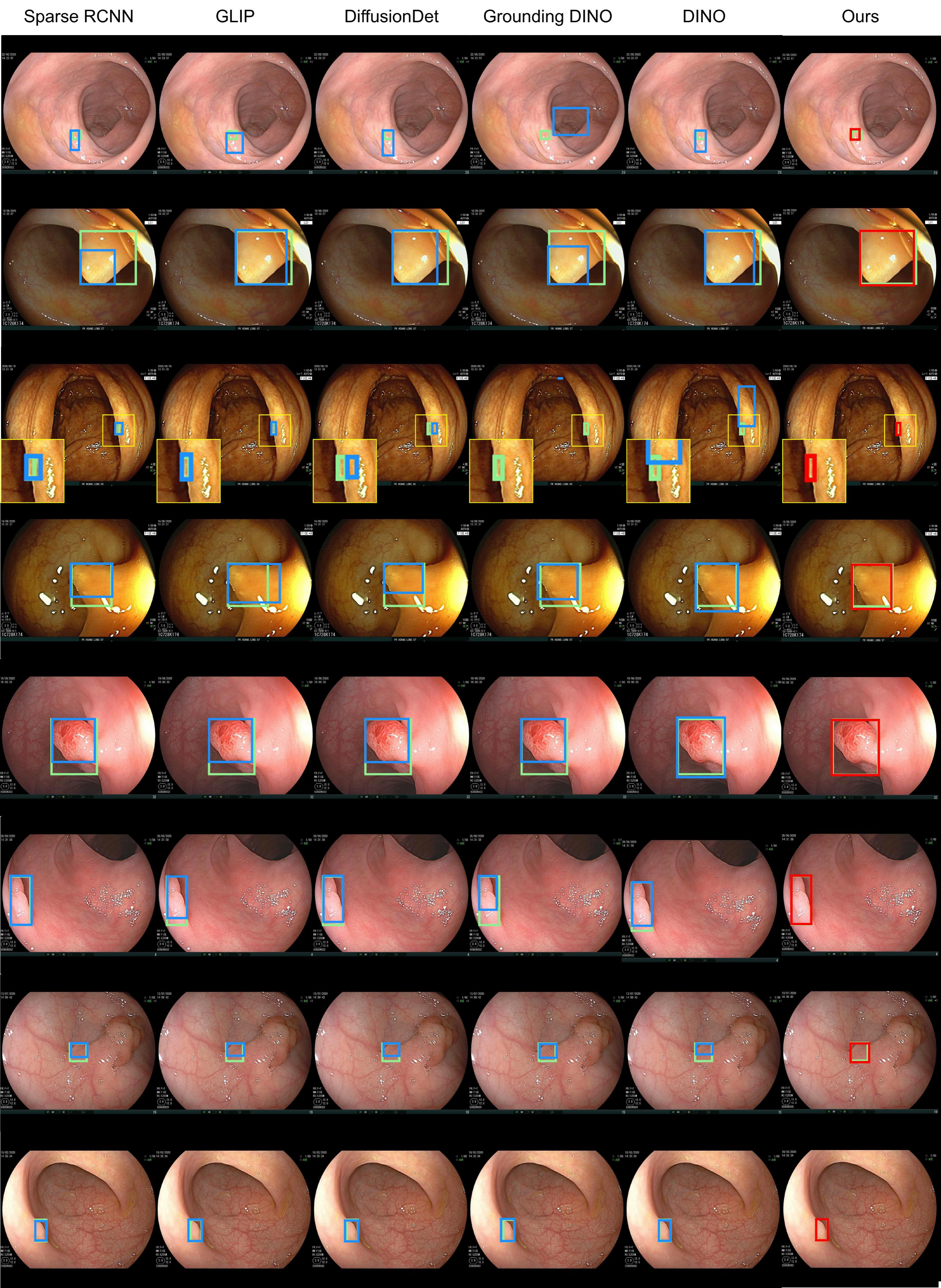}
    \caption{
        \textbf{Qualitative comparisons on \uline{Colonoscopy} images.}
        \textcolor{blue}{Blue} boxes indicate predictions from baseline detectors 
        (Sparse R-CNN, GLIP, DiffusionDet, Grounding DINO, DINO), 
        \textcolor{green!60!black}{Green} boxes denote ground-truth annotations, 
        and \textcolor{red}{Red} boxes show predictions from our method.
    }
    \label{fig:colonoscopy_qualitative}
\end{figure*}

\begin{figure*}[t]
    \centering
    \includegraphics[width=.9\linewidth]{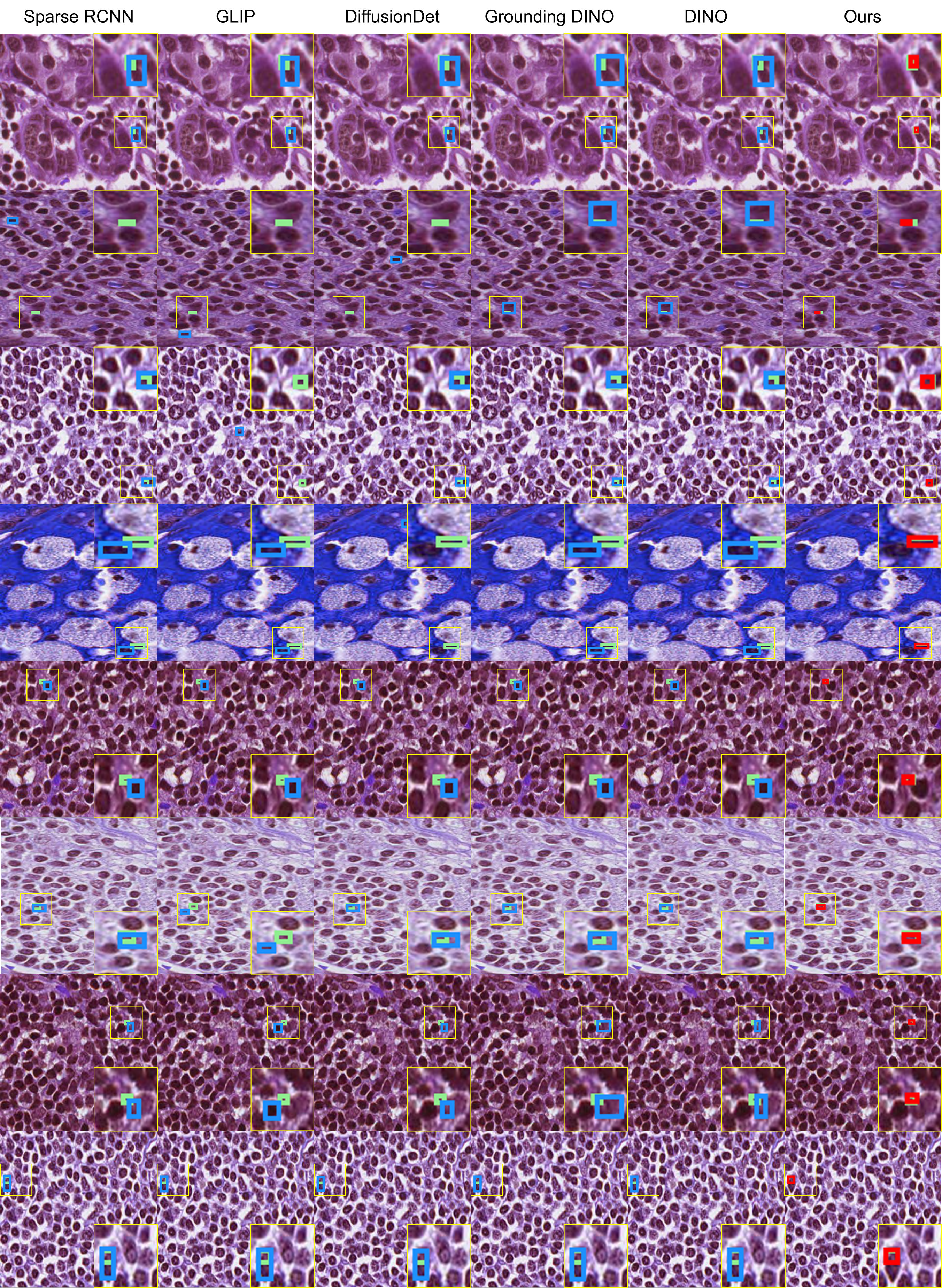}
    \caption{
        \textbf{Qualitative comparisons on \uline{Pathology} images.}
        \textcolor{blue}{Blue} boxes indicate predictions from baseline detectors 
        (Sparse R-CNN, GLIP, DiffusionDet, Grounding DINO, DINO), 
        \textcolor{green!60!black}{Green} boxes denote ground-truth annotations, 
        and \textcolor{red}{Red} boxes show predictions from our method.
    }
    \label{fig:pathology_qualitative}
\end{figure*}

\begin{figure*}[t]
    \centering
    \includegraphics[width=.9\linewidth]{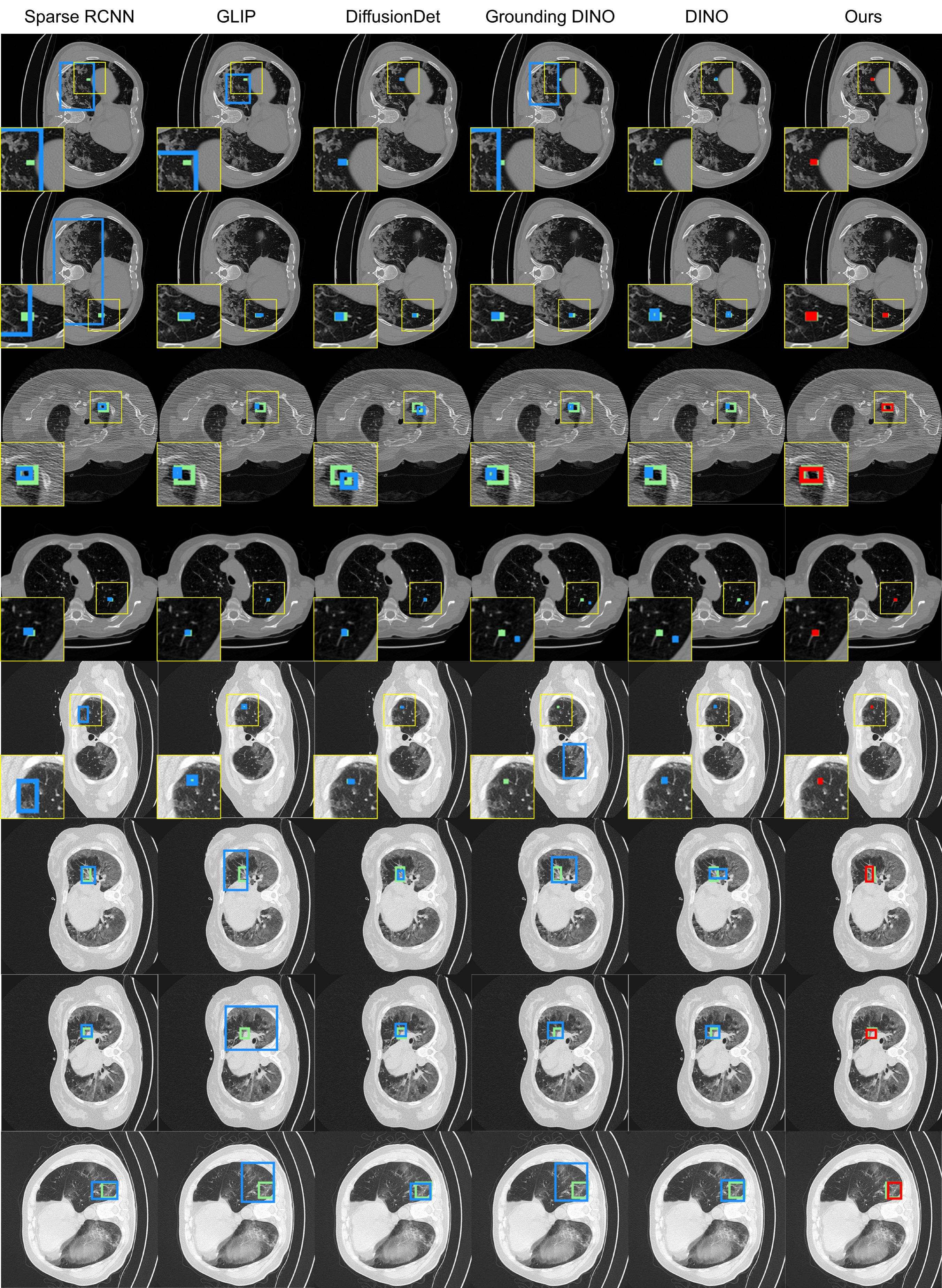}
    \caption{
        \textbf{Qualitative comparisons on \uline{CT} images.}
        \textcolor{blue}{Blue} boxes indicate predictions from baseline detectors 
        (Sparse R-CNN, GLIP, DiffusionDet, Grounding DINO, DINO), 
        \textcolor{green!60!black}{Green} boxes denote ground-truth annotations, 
        and \textcolor{red}{Red} boxes show predictions from our method.
    }
    \label{fig:ct_qualitative}
\end{figure*}

We present additional qualitative results for each modality 
(\uline{CXR}, \uline{MRI}, \uline{Colonoscopy}, \uline{CT}, and \uline{Pathology}) 
in Figures~\ref{fig:cxr_qualitative}–\ref{fig:ct_qualitative}. 
Each figure compares baseline detectors (Sparse R-CNN, GLIP, DiffusionDet, 
Grounding DINO, DINO) with our method, with 
\textcolor{blue}{blue} boxes for baselines, 
\textcolor{green!60!black}{green} boxes for ground truth, 
and \textcolor{red}{red} boxes for our predictions.

\end{document}